\pdfoutput=1
\documentclass[10pt,journal,compsoc]{IEEEtran}
\usepackage{amsmath}
\usepackage{graphicx}
\usepackage{amssymb}
\usepackage{multirow}
\usepackage{multicol}
\usepackage{enumitem}
\usepackage{color}
\usepackage{url}
\usepackage{array}
\usepackage{stfloats}
\usepackage{siunitx}[group-separator={,}]
\usepackage{cases}
\usepackage{soul}

\ifCLASSOPTIONcompsoc
  \usepackage[nocompress]{cite}
\else
  \usepackage{cite}
\fi

\ifCLASSOPTIONcompsoc
 \usepackage[caption=false,font=footnotesize,labelfont=sf,textfont=sf]{subfig}
\else
 \usepackage[caption=false,font=footnotesize]{subfig}
\fi

\newcommand{\note}[2]{{#2}}
\newcommand{\fix}[1]{{}}

\newcommand{\oursbase}{Ours$/_{base}$}
\newcommand{\oursbaseBplus}{Ours$/_{base}^{\gamma+}$}
\newcommand{\oursbaselarge}{Ours$/_{base}^{L}$}
\newcommand{\oursbaselargeBplus}{Ours$/_{base}^{L\gamma+}$}

\begin{document}
\title{Taming Reversible Halftoning\\via Predictive Luminance}

\author{
        Cheuk-Kit~Lau, 
        Menghan~Xia, 
        Tien-Tsin~Wong

        \IEEEcompsocitemizethanks{
        \IEEEcompsocthanksitem Cheuk-Kit Lau and Tien-Tsin Wong are with the Department of Computer Science and Engineering, The Chinese University of Hong Kong.\\
        \protect~E-mail:\{cklau21,ttwong\}@cse.cuhk.edu.hk
        \IEEEcompsocthanksitem Menghan Xia is with Tencent AI Lab.\\
        \protect~E-mail:menghanxyz@gmail.com
        \IEEEcompsocthanksitem This project is partially funded by RGC Direct Grant (CUHK Project Code \#4055152).
        }
}

\markboth{}
{Lau \MakeLowercase{\textit{et al.}}:Taming Reversible Halftoning\\via Predictive Luminance}


\IEEEtitleabstractindextext{%
\begin{abstract}

Traditional halftoning usually drops colors when dithering images with binary dots, which makes it difficult to recover the original color information.
We proposed a novel halftoning technique that converts a color image into a binary halftone with full restorability to its original version.
Our novel base halftoning technique consists of two convolutional neural networks (CNNs) to produce the reversible halftone patterns, and a noise incentive block (NIB) to mitigate the flatness degradation issue of CNNs.
Furthermore, to tackle the conflicts between the blue-noise quality and restoration accuracy in our novel base method, we proposed a predictor-embedded approach to offload predictable information from the network, which in our case is the luminance information resembling from the halftone pattern.
Such an approach allows the network to gain more flexibility to produce halftones with better blue-noise quality without compromising the restoration quality.
Detailed studies on the multiple-stage training method and loss weightings have been conducted.
We have compared our predictor-embedded method and our novel method regarding spectrum analysis on halftone, halftone accuracy, restoration accuracy, and the data embedding studies.
Our entropy evaluation evidences our halftone contains less encoding information than our novel base method.
The experiments show our predictor-embedded method gains more flexibility to improve the blue-noise quality of halftones and maintains a comparable restoration quality with a higher tolerance for disturbances.

\end{abstract}

\begin{IEEEkeywords}
Reversible halftoning, deep learning, blue-noise.
\end{IEEEkeywords}}

\maketitle

%
\IEEEpeerreviewmaketitle

\IEEEraisesectionheading{\section{Introduction}
\label{sec:introduction}}

\IEEEPARstart{H}{alftoning} is commonly used in the printing industry\cite{ulichney1988dithering} to reproduce tone with limited colors, e.g. black
and white, due to the cost consideration. 
The original image's color and fine details are inevitably lost during this process. 
This makes the originals nearly impossible to be recovered from these degraded halftones. 
Even the state-of-the-art inverse halftoning methods\cite{xia2018deep, kim2018deep} can only recover an approximate grayscale version since the color is usually dropped before halftoning. 
Apparently, resolving this dilemma requires a fore-looking halftoning technique that retains the necessary information for restoration. 
In this paper, we conducted a thorough study to explore this problem.

\begin{figure}[!t]
    \centering

    \includegraphics[width=0.48\textwidth]{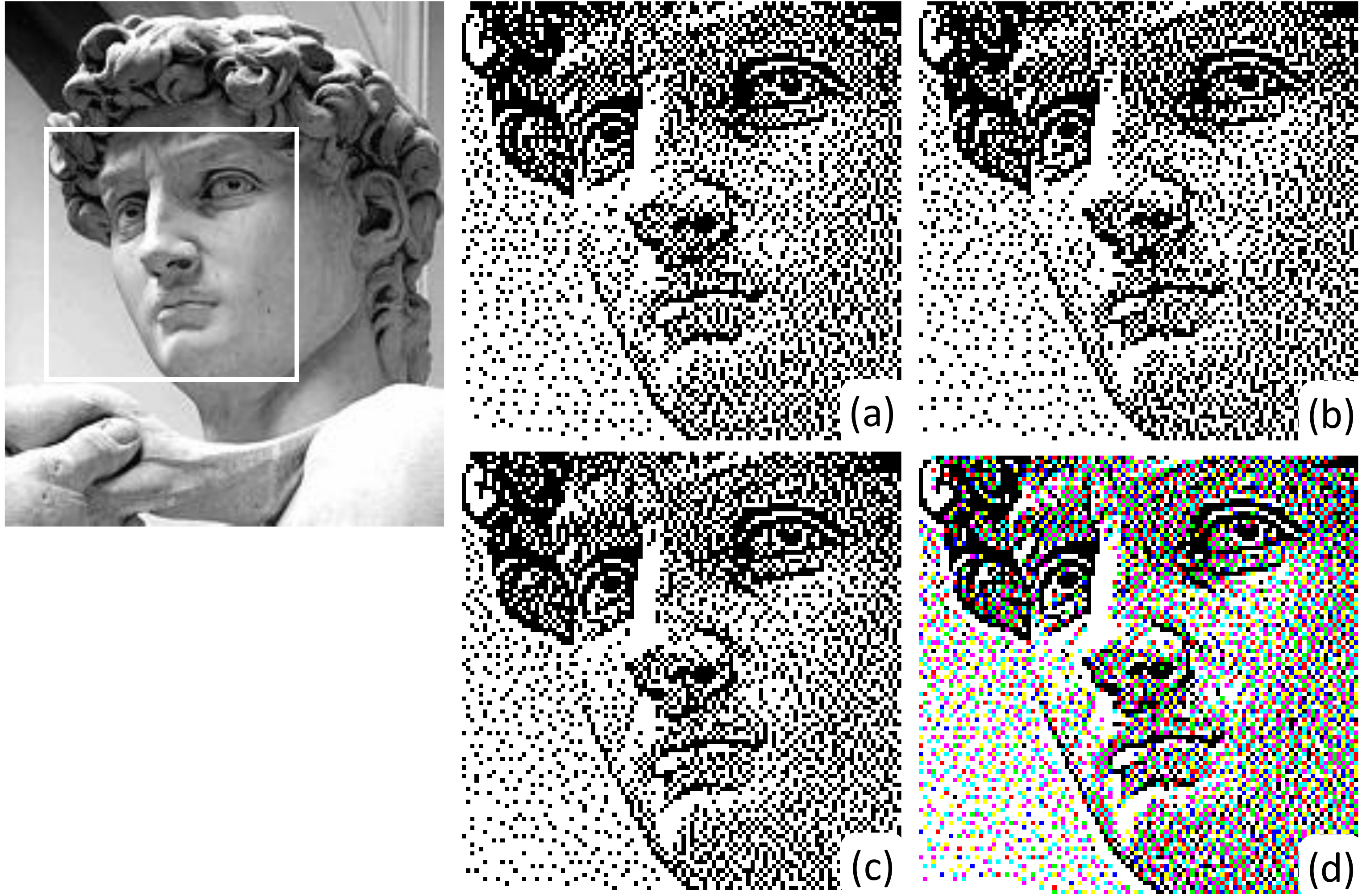}
    \caption{ Observation: the halftone variants of (a) (b) (c) present similar visual quality but with different binary patterns, as the overlaid RGB image visualized in (d). It shows the possibility of modulating the patterns for additional usage.}
    \label{fig:observation}
\end{figure}

Traditional halftoning methods distribute halftone dots mainly for tone reproduction. 
We observe that this target still permits certain perturbation in terms of the desired binary pattern, as evidenced in Fig.~\ref{fig:observation}. 
It indicates the possibility of utilizing such a degree of freedom for additional usage, i.e., embedding the potentially missing color information and fine details. 
Formally, this brings out a new concept, i.e., reversible halftoning, which converts a color image to a halftone that possesses restoration ability to the original color version. 
Inspired by invertible grayscale\cite{xia2018invertible}, we adopt the invertible generative model to formulate our problem. 
However, generating quality halftones is more challenging than decolorization. 
The challenges lie in the flatness degradation of CNNs in halftoning and the difficulty in achieving vivid visual simulation and accurate information embedding with 1-bit pixels. To address flatness degradation, we propose a Noise Incentive Block (NIB) that introduces spatial variation to the feature space while reserving the information intactness. To achieve the binary halftone, we propose a binary gate that takes gradient propagation tricks to allow training with quantization.

Anyhow, as reported in our preliminary study~\cite{xia2021deep}, the binary encoding space is limited and causes sacrifice for the blue-noise property against the restoration accuracy. Inspired by the predictive coding concept~\cite{said1993reverisble}, we promote the encoding framework by exploiting the predictive power from the inverse halftone module. The intuition is that most luminance information could be inferred from the halftone, and removing luminance information from the encoding stage offers more capacity for blue-noise realization.
The model is trained end-to-end with highly mixed objectives, formulated as three loss terms: halftone loss, restoration loss, and luminance loss.
Particularly, we propose a guiding-ware training scheme to circumvent the tricky converging issue of multi-objective optimization.

Extensive evaluation and ablation study demonstrate that the proposed predictive encoding model allows a good balance among the visual simulation, blue-noise profile, and restoration accuracy for reversible halftoning. The trained model achieves very competitive performance against traditional halftoning algorithms in halftoning quality while still maintaining decent restoration accuracy of the original color image.


The preliminary version of this manuscript presented two distinct contributions. Firstly, we introduced a novel method for reversible halftoning that enhances the functionality of existing halftoning applications. This method circumvents the ill-posed inverse halftoning problem at its source. Secondly, we proposed a model-agnostic plug-in, the noise incentive block, which effectively addresses the flatness degradation of CNN.

In this manuscript, our primary focus is to promote the invertible generation framework with a predictive coding concept. Our objective is to demonstrate the efficacy of this framework in reducing the encoding burden and improving the embedded halftone quality.

\section{Background}
\label{sec:background}

\subsection{Image Halftoning}
\label{subsec:halftoning}
Digital halftoning has been widely studied over the past decades. 
The goal is to render images in only two levels of pixel values, black and white. 
It creates an illusion of the continuous tone of the original image through the spatial distances between black and white dots. 
Traditional deterministic approaches include 
    ordered dithering\cite{limb1969design, lippel1971effect, bayer1973optimum}, 
    error diffusion\cite{floyd1976adaptive,ostromoukhov2001simple, ostromoukhov2004fast}, 
    dot diffusion\cite{knuth1987digital}, 
    and direct binary search\cite{chandu2013direct}.
They aim to produce halftone images that preserve the local tone of the original image while with minimal artifacts.

Since humans are perceptually more aware of artifacts in low-frequency areas, an ideal halftone image should contain the blue-noise property.
The blue-noise property corresponds to visually pleasing\cite{ulichney1988dithering} and minimal low-frequency components\cite{mitsa1992digital}.
There are several works to achieve this, such as using
    perturbed error diffusion\cite{ulichney1988dithering},
    blue-noise mask\cite{mitsa1992digital, yu1997adaptive},
    diffusion parameter set optimization\cite{ostromoukhov2001simple, zhou2003improving}, 
    and tile-based methods\cite{lagae2008tile}.

Although focusing on blue-noise rendering can produce a smooth and evenly distributed surface, fine details such as edges and complex structures will be blurred. 
Many proposed works aim to improve the halftone images using edge enhancement\cite{roetling1976halftone, eschbach1991error,lai2003algorithms, li2006edge}. 
Pang et al.\cite{pang2008structure} first introduced structural similarity and tonal similarity into the optimization function, followed by Chang et al.\cite{chang2009structure} optimized the error diffusion algorithm with structural similarity.

Some neural-network-based approaches\cite{anastassiou1988neural, crounse1993image} aim to produce halftone images in a deterministic manner.

\subsection{Inverse Halftoning}
\label{subsec:inverse_halftoning}

In the early printing industry, many images in newspapers, magazines, and books are halftone printings.
``Inverse halftone'' dedicates to restoring the continuous tone of images from the halftone images.
It is an ill-posed problem because the fine details have been lost in the halftoning process.
The simplest method is to process the halftone image with a low-pass filter\cite{kim1995inverse, wong1995inverse, chen1997adaptive}.
However, such a method will also remove edge information.
Kite et al.\cite{kite1998high} proposed a kernel function built from local gradients to preserve high-frequency details.
Xiong et al.\cite{xiong1999inverse} proposed to extract edge information and discard background noise via wavelet decomposition.
Some works reformulate the continuous-tone restoration problem as a projection onto convex sets (POCS)\cite{hein1993halftone, unal2001restoration}.
Ting and Riskin\cite{ting1994error} proposed using a look-up table (LUT) to obtain a temporary grayscale image. 
Mese and Vaidyanathan\cite{mese2001look} further proposed restoring the grayscale image using LUT without any linear filtering techniques.
Both approaches improve the efficiency of restoring continuous-tone images.
Therefore, many dictionary learning-based approaches have been proposed since then\cite{lee2009reversible, lee2010new, son2012inverse, son2014local, freitas2016enhancing, zhang2018sparsity}. 

Yue and Chen\cite{yue1995auto} proposed using Hopfield neural network\cite{hopfield1982neural} based optimization model to inverse halftoning.
Huang et al.\cite{huang2008neural} proposed using a radial basis function neural network to restore the continuous tone from the halftone input.
However, the quality of inverse halftoning is highly dependent on the starting halftone method.

Recently, deep learning approaches have been explored by authors.
Xiao et al.\cite{xiao2017deep} and Gao et al.\cite{gao2019deep} proposed inverse halftone via U-Net structure with convolution layers.
Xia and Wong\cite{xia2018deep} improved the restoration quality by introducing residual learning layers to predict enhanced details further.
Kim and Park\cite{kim2018deep} proposed a generative adversarial network (GAN) with object categories prediction and edge information extraction.
Besides restoring grayscale images, restoring color from halftone images is harder. 
It is because more information is needed to fill-ins instead of luminance only.
Yen et al.\cite{yen2021inverse} restored color images by concatenating the inverse halftone and colorization stages. Such a method requires extra information to hint at the network to predict color from the intermediate grayscale image.

\subsection{Reversible Generation}                
\label{subsect:reverible_generation}

The reversible generation topic has been widely studied in the data hiding field.
Major tasks applications include hiding watermarks or copyright declarations in images \cite{luo2014disparity, yang2015reversible, zhu2018hidden}.
Also, authors have explored methods to hide the color information in the grayscale version image.
Queiroz and Braun\cite{de2006color} proposed hiding chrominance channels into subbands from wavelet transform.
Xu and Chan\cite{xu2017improving} proposed hiding the chrominance channels specifically in high-frequency areas of the grayscale version via error diffusion techniques.

Recently, CNNs have gained massive success in image processing tasks.
By considering the grayscale image as a latent representation of the color image, Xia et al.\cite{xia2018invertible} proposed an encoding-and-decoding framework to generate reversible grayscale images that can be reversed back to their color version.
Ye et al.\cite{ye2020invertible} further proposed using the dual features extractions to improve the restoration quality.
A similar framework is adopted in other tasks, such as image resampling\cite{li2018learning, xing2022scale} and image retargeting\cite{tan2019cycle}.
Another approach, invertible neural networks (INNs)\cite{dinh2016density, jacobsen2018revnet, kingma2018glow, behrmann2019invertible, ardizzone2019guided, xiao2020invertible, liu2021invertible, zhao2021invertible}, generates latent representation without loss of information; however, it relies on explicitly structured network architecture. Such constraint generally makes the training tricky and unstable.
In our preliminary study~\cite{xia2021deep}, we adopted the invertible generation model as~\cite{xia2018invertible}, and the limited encoding space of binary pattern causes the trade-off between the blue-noise quality and the restoration quality. In this paper, we promote the encoding framework with a predictive coding concept, i.e., removing the luminance information from the encoding stage and inferring it from the halftone pattern, which facilitates making a practically better balance between the visual quality and data embedding accuracy.

\section{Reversible Halftoning}
\label{sec:method}

We aim to learn reversible binary patterns toward halftoning color images, which is required to offer visual pleasantness and embed restoration-necessary information in the meantime. 
The key idea is to encode the color information into the halftone image and restore the color image by decoding the halftone image.
We first adopted the autoencoder design, where the latent feature is represented in the halftone patterns, to approach the problem.
However, the halftone patterns have to fulfill certain objectives:
1) the distribution of dots should resemble the continuous tone of its grayscale version perceptually;
2) the distribution of dots should maintain high blue-noise quality;
and 3) the color information should be embedded into the distribution of dots.
This poses a challenge to the novel autoencoder approach because the latent feature is not just a representation of the embedded information, but for fulfilling all three objectives simultaneously.

\begin{figure*}[!t]
\centering

\includegraphics[width=\textwidth]{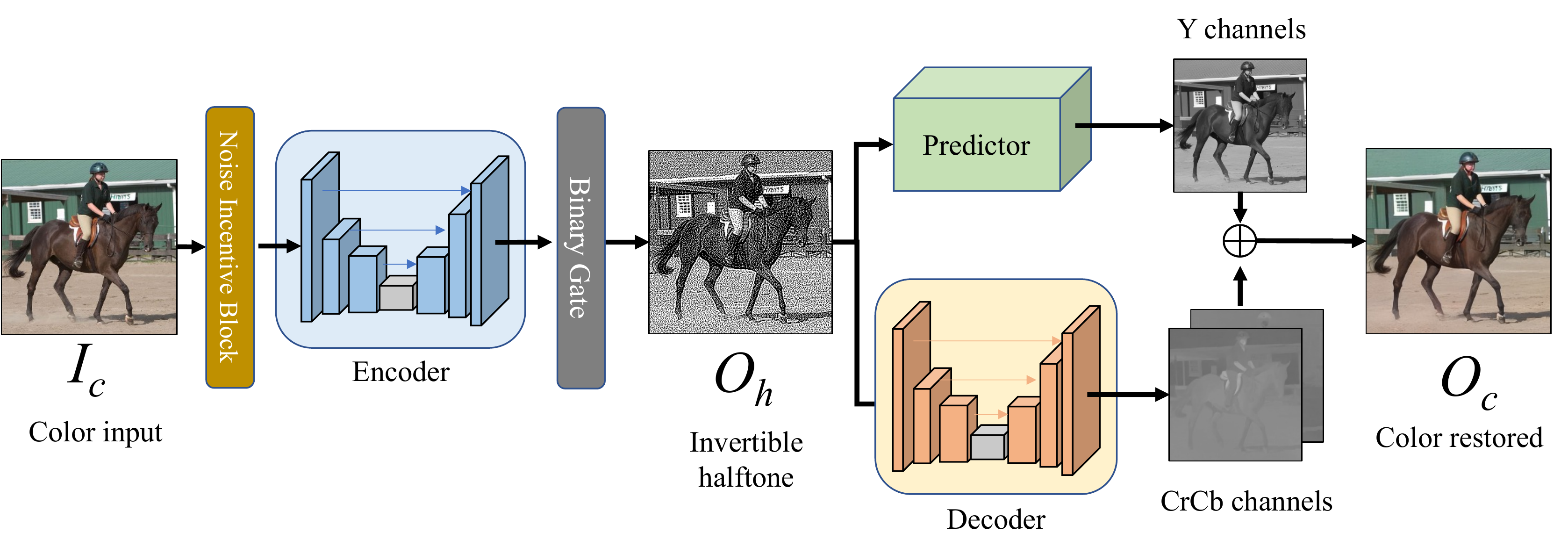}
\caption{Overview of our network architecture with embedded luminance predictor. $\oplus$ denotes the concatenation operation.}
\label{fig:overview}
\end{figure*}

\begin{figure}[!t]
\centering
\includegraphics[width=\linewidth]{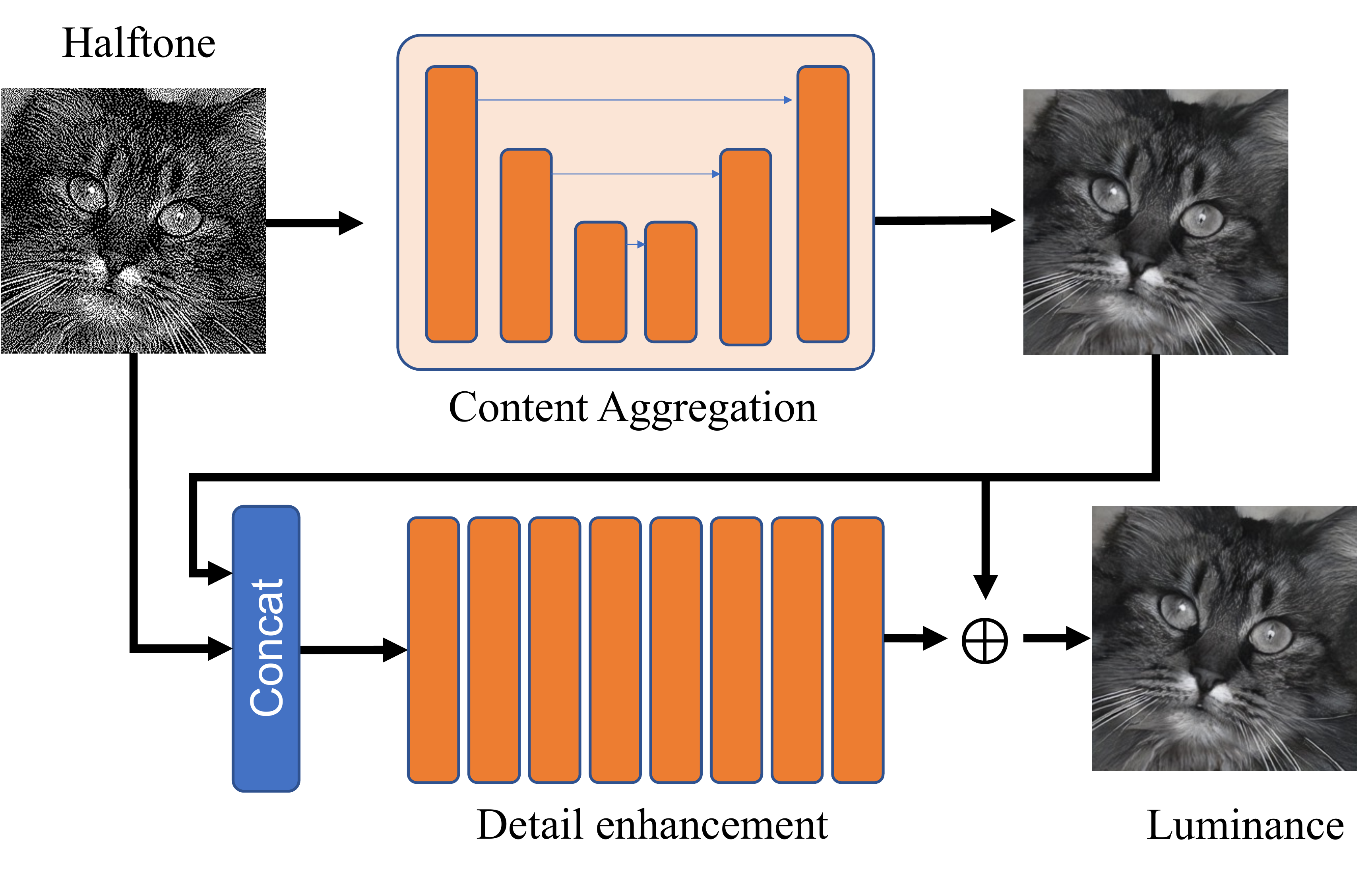}
\caption{Overview of the predictor architecture \cite{xia2018deep}. $\oplus$ denotes the addition operation.}
\label{fig:predictor-overview}
\end{figure}

\subsection{Embedding Framework with Predictive Luminance}
\label{subsec:predictive_encoding}

\noindent \textbf{Concept of Predictive Coding.} \quad
The concept of predictive coding had been described in different areas. 
In neuroscience, "predictive coding" suggests that the brain solves inverse problems via an internal model of the world\cite{spratling2017review, spratling2008predictive}. 
It provides an explanation of how our brain receives and reduces redundant signals.
Such an idea was also established in the signal-processing domain.
The key idea is to compress data with discarded information and restore the data by predicting the discarded information back.
Predictable information shall be excluded from the compressed data.
The compressed data should only include the residual error between the predicted and the actual values.
Such an approach significantly increases the compression ratio.
Predictive coding appears in various applications, such as image compression\cite{weinberger2000loco}, temporal video compression\cite{wiegand2003overview} and representation learning\cite{oord2018representation, han2019video}. 

Our problem is similar to the data compression settings, where information is compressed (encoded) and restored (decoded).
Our novel autoencoder method suffers the drawback of encoding information into the halftone pattern.
Since we train the network to encode and decode information in RGB space, the encoder will encode all information in RGB as it can.
However, due to the binary level of pixels and the halftone image having to resemble the continuous tone of its input, the encoding space available for encoding is further limited.
In our base method, blue-noise quality has been sacrificed.
If we remove some information from the limited encoding space but put it back in the restoration stage, the network should have more freedom to produce halftone patterns while maintaining its restoration ability.
On the other hand, we know the work of inverse halftone has been long studied and well-developed.
State-of-the-art work\cite{xia2018deep} can predict the continuous tone from halftone images with fine details.
We can offload the luminance information from the encoding-decoding pipeline, thus constraining the network to sample the subspace of chrominance only.
In the restoration stage, we extend the network with a predictor module, an inverse halftone module, to restore the offloaded luminance information.
In this manuscript, we aim to improve the blue-noise quality of our halftone image through this spared encoding space.

We extend the design established in \oursbase \cite{xia2021deep}.
Our network consists of three main components(\figurename~\ref{fig:overview}):
\setitemize[1]{itemsep=1pt,partopsep=0pt,parsep=\parskip,topsep=3pt}
\begin{itemize}
    \item An encoder that encodes color information into the generated halftone image;
    \item A predictor that predicts the luminance channel from the encoded halftone image;
    \item A decoder that restores the chrominance channels from the encoded halftone image.
\end{itemize}

Given an RGB image $I_c$, we construct a reversible halftone image $O_h$ by the encoder $\mathrm{E}$ and the binary gate $\mathrm{B}$:
\begin{align} 
    \Tilde{O}_h &= \mathrm{E}(\ \mathrm{Nib}(I_c)\ ) \\
    O_h &= \mathrm{B}( \Tilde{O}_h ) \label{eq:encoder}
\end{align}
The details of the noise incentive block $\mathrm{Nib}(\cdot)$ is discussed in section \ref{subsec:incentive_block}.
The encoder generates a \emph{pseudo halftone image} $\Tilde{O}_h$, in which each pixel are real value ranging from 0 to 1.
The binary gate quantizes the pixels in the pseudo halftone image from real value to either 0 or 1.
Then we feed $O_h$ from \eqref{eq:encoder} into two networks,
a decoder network $\mathrm{D}$ to restore the chrominance channels $O_c^{ch}$, and a predictor network $\mathrm{P}$ to predict the luminance channel $O_c^l$.
\begin{align}
    O_c^{ch} &= \mathrm{D}(O_h) \\
    O_c^l &= \mathrm{P}(O_h) 
\end{align}
Finally, we obtain the restored color image $O_c$ by concatenating those three channels and convert to RGB color space.
Our color space conversion function follows the standard specified in \cite{bt2011studio}.

\subsection{Network Architecture}
\label{subsec:architecture}

We adopt the U-shaped architecture for both the encoder and decoder networks. 
Both networks share a similar structure, containing three downscale blocks, three upscale blocks, four residual blocks, and two convolution blocks. 
We adopt U-Net as the network backbone because of its enlarged receptive field, and other qualified CNN architectures may also work. 
We adopted the \cite{xia2018deep} model as our predictor module.
Any other inverse halftone module may also work.
Additionally, we propose two special designs within this network: 
the noise incentive block to mitigate the flatness degradation introduced by CNN; 
and the binary gate to encourage the network to generate near-binary pixels.
The base architecture, which does not include the predictor module, is denoted as \oursbase~in this manuscript.

\subsubsection{Noise Incentive Block}
\label{subsec:incentive_block}

We uncovered a phenomenon that we refer to as "flatness degradation," which arises from the convolutional paradigm with spatially shared kernels when presented with flat inputs. This phenomenon leads to a scaling operation that applies the same parameters across the input and produces a constant signal, thereby impeding the ability of CNNs to dither a constant grayness. This, in turn, hinders the formulation of the blue-noise profile, which is primarily measured over the constant grayness. To address this issue, we propose the Noise Incentive Block (NIB), which introduces spatial variation to the feature representation while preserving the original input. By preprocessing the color image before passing it to the encoder, our dithering network is able to generate binary halftone in flat regions. The NIB also enables us to formulate the blue-noise profile through low-frequency constraints on dithered constant-grayness. The example of NIB-equipped results is located in \figurename~\ref{fig:degradation} in the supplementary.

\subsubsection{Binary Gate}
\label{subsec:binary_gate}

Another special design for the dithering network is the binary gate $\mathrm{B}(\cdot)$ that quantizes the network output $\Tilde{O}_h$ to be a strict binary image $O_h = \mathrm{B}(\Tilde{O}_h)$. 
We explicitly adopt a binary gate because the soft non-binary penalty is insensitive to tiny deviations, i.e., near-0 or near-1 valued pixels, which is vulnerable to quantization when stored as a 1-bit bitmap and thus hurts the restoration accuracy. However, one obstacle should be noted: the binary gate is non-differentiable. 
To enable the joint training, we use Straight-Through Estimator \cite{bengio2013estimating} on the binarization when calculating the gradients.

\subsubsection{Predictor}

\figurename~\ref{fig:predictor-overview} shows an overview of the predictor module.
We notice that halftone patterns inherently convey luminance and structural information, regardless of whether they are encoded with color information or not. As a result, we have implemented Xia's\cite{xia2018deep} inverse halftone module to predict continuous luminance information, thereby allowing us to concentrate on encoding chrominance information. The predictor consists of two key components: the content aggregation block, which incorporates three downscale blocks, three upscale blocks, and four residual blocks; and the detail enhancement block, which employs eight residual blocks to improve the predicted luminance details.

\subsection{Loss Function}
\label{subsec:loss}

We trained our network with the following loss functions:
the halftone loss $\mathcal{L}_{half}$; the restoration loss $\mathcal{L}_{restore}$; and the luminance loss $\mathcal{L}_{lumin}$.
We trained our network in multiple stages; the detailed combination of loss functions and their corresponding coefficients are discussed in Section \ref{sec:method.training_strategy}.

\subsubsection{Halftoning}
\label{sec:method.loss.halftoning}

We adopted the halftone loss $\mathcal{L}_{half}$ to train the network to generate the desired reversible halftone image. Our halftone loss is formulated as:
\begin{equation}
    \mathcal{L}_{half} = \alpha \cdot \mathcal{L}_{bin} + \beta \cdot \mathcal{L}_{tone} + \gamma \cdot \mathcal{L}_{blue}
\end{equation}
where $\mathcal{L}_{bin}$ denotes the binary loss; $\mathcal{L}_{tone}$ denotes the tone loss; and $\mathcal{L}_{blue}$ denotes the blue-noise loss.

Let $\tilde{O}_h$ be the \emph{pseudo halftone image} generated by the encoder $\mathrm{E}$ but before the quantization layer $\mathrm{Q}$.
Since quantization on $\tilde{O}_h$ cannot be differentiated, binarization loss takes a crucial role in encouraging the network to produce binary intensity values on the halftone image.
It is formulated as
\begin{equation}
    \mathcal{L}_{bin} = || \mathrm{B}(\tilde{O}_h) - \tilde{O}_h ||_1
\end{equation}
where $|| \cdot ||_1$ denotes the $L_1$ norm.
$\mathrm{B}(\cdot)$ denotes the binary gate.

Based on the tone similarity concept, which was proposed by \cite{pang2008structure}, we applied tone loss $\mathcal{L}_{tone}$ to encourage the halftone image $O_h$ to resemble the tone of the input image. It is formulated as follows:
\begin{equation}
    \mathcal{L}_{tone} = || G(I_{gray}) - G(O_h) ||_2
\end{equation}
where $G(\cdot)$ denotes a Gaussian filter with kernel size $11 \times 11$ and sigma $2.0$; 
$I_{gray}$ denotes the grayscale version of color input $I_c$;
$|| \cdot ||_2$ denotes the $L_2$ norm.

To train the network to produce halftone with blue-noise property, we adopted the blue-noise loss $\mathcal{L}_{blue}$ suggested by \cite{xia2021deep}. 
Its basic idea is to restrict the network to generate minimal low-frequency components because they are more noticeable to the human eye. 
Therefore, we prepared a set of plain-color images $\mathcal{P}$. 
For each training iteration, after the color image from the dataset has been passed to the network, we randomly draw a plain color image $p \in \mathcal{P}$. 
A halftone image $z_p$ is obtained by passing $p$ into the network.
The blue-noise loss is formulated as
\begin{equation}
    \mathcal{L}_{blue} = || [DCT(z_p) - DCT(p_{gray})] \odot M ||_2
\end{equation}
where $p_{gray}$ denotes the grayscale version of $p$. 
$DCT(\cdot)$ denotes the discrete cosine transformation function.
$M$ denotes the binary mask. 
We set $M$ to only allow the first 3.8\% of low-frequency DCT coefficients to pass through. 

Compared to our preliminary version\cite{xia2021deep} of this manuscript, we dropped the structure loss suggested by \cite{pang2008structure} since it has no significant effect on our training outcome.

\subsubsection{Restoration}
\label{sec:method.loss.restoration}

We constructed the restoration loss as
\begin{equation}
     \mathcal{L}_{restore} = \zeta \cdot \mathcal{L}_{chromin}
                            + \eta \cdot \mathcal{L}_{percep}
\end{equation}
where $\mathcal{L}_{chromin}$ denotes the chrominance loss; and $\mathcal{L}_{percep}$ denotes the perceptual loss.
The chrominance loss trains the decoder to extract chrominance information from the encoded halftone image.
Given a restored chrominance channels $O_c^{ch}$, the chrominance loss are formulated as
\begin{equation}
    \mathcal{L}_{chromin} = || I_c^{ch} - O_c^{ch} ||_2
\end{equation}

The perceptual loss trains the network to resemble color signals at the perceptual level.
We adopted the perceptual loss $\mathcal{L}_{percep}$ suggested by \cite{xia2021deep}, which is formulated as
\begin{equation}
    \mathcal{L}_{percep} = || \Psi(I_c) - \Psi(O_c) ||_2
\end{equation}
where we denote $\Psi(\cdot)$ as the latent feature extracted from the \emph{conv4\_4} layer of the pre-trained VGG-19 module\cite{simonyan2014very}.

The luminance loss trains the predictor to generate a continuous luminance channel from the halftone image. 
Since we adopted the inverse halftone module from \cite{xia2018deep}. We take the full loss function of \cite{xia2018deep} as our luminance loss, 
$\mathcal{L}_{lumin}$.
\begin{equation}
    \mathcal{L}_{content} = || \hat{O}^l_c - I_{gray} ||_2
\end{equation}
\begin{equation}
    \mathcal{L}_{full} = w_a || \Psi(O^l_c) - \Psi(I_{gray}) ||_2 + || O^l_c - I_{gray} ||_1
\end{equation}
\begin{equation}
    \mathcal{L}_{lumin} = \mathcal{L}_{content} + w_b \mathcal{L}_{full} \label{eq:lumin_loss}
\end{equation}
where $\hat{O}^l_c$ denotes the initial predicted grayscale image from the content aggregation module in \cite{xia2018deep}.
We set the coefficients to the default value stated in \cite{xia2018deep}, where $w_a = 2.0 \times 10^{-6}$, $w_b = 1.5$.

\subsection{Training Strategy} 
\label{sec:method.training_strategy}

\begin{figure}[!t]
    \centering
    \includegraphics[width=\linewidth]{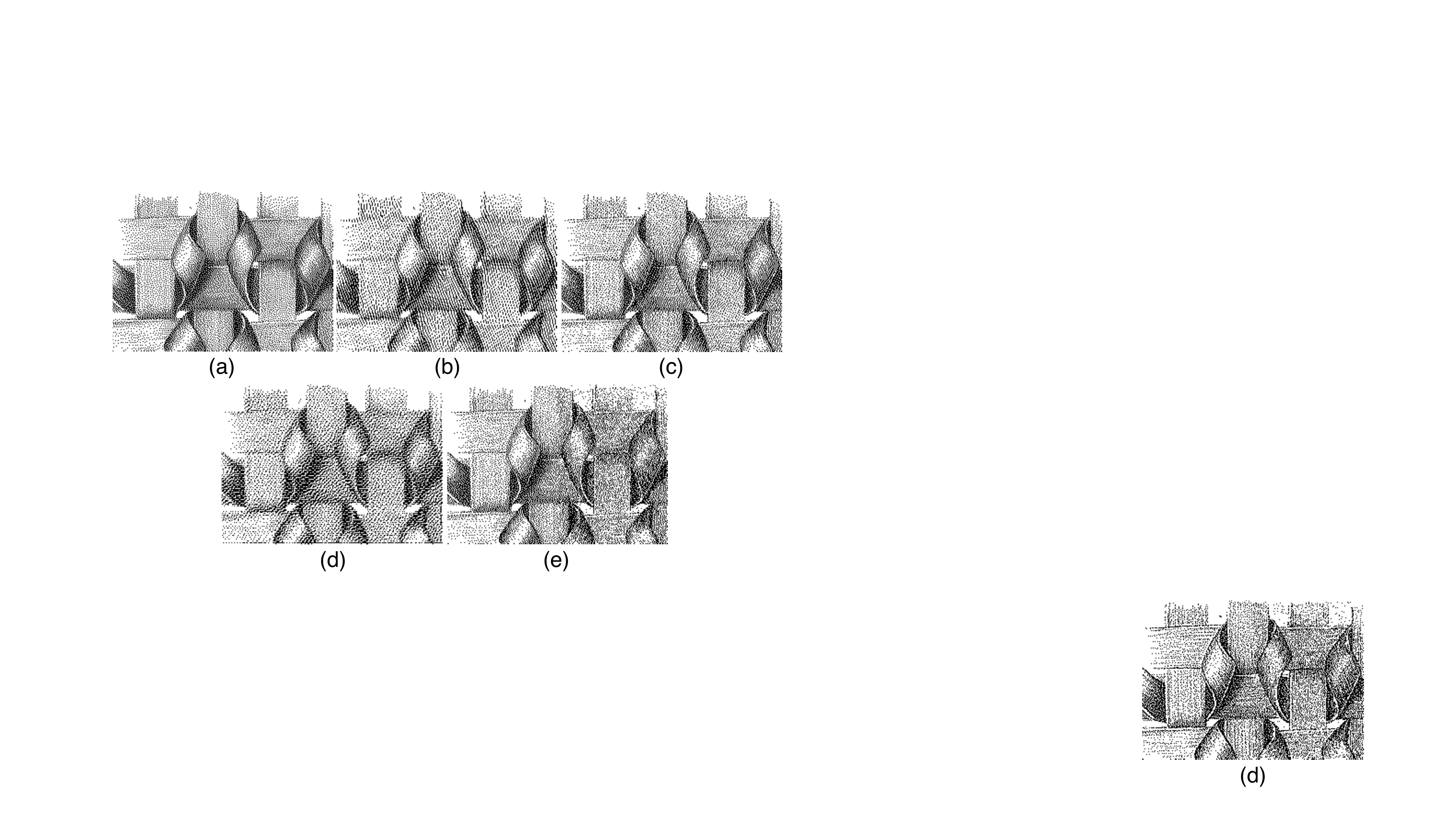}
    \caption{
    Halftone generated by models trained with and without guidance loss in different stages.
    (a) Error diffusion;
    (b) warm-up training for 130 epochs w/o guidance loss;
    (c) warm-up training for 28 epochs with guidance loss; 
    (d) our stage two w/o guidance loss; 
    (e) our stage two with guidance loss.
    }
    \label{fig:guidance_loss}
\end{figure}

Training the whole model from scratch is vulnerable to a local minimum because of the challenging optimization target. 
To circumvent this problem, we propose to adopt a warm-up training scheme. 
In the first stage, we aim to warm up the dithering network alone, so that it can generate visually pleasant halftone images. 
To stabilize the training, the binary gate is temporally removed. 
Unfortunately, this relaxation still fails to guarantee satisfactory halftones in \figurename~\ref{fig:guidance_loss}(b), and it is even associated with slow convergence, \note{loss_curve}{as shown in \figurename~\ref{fig:loss-curve}(green curve) in the supplementary.} 
To boost the training, we propose explicitly providing a reference halftone image $I_h$ to guide the training. 
For simplicity, the classical error diffusion \cite{ostromoukhov2001simple} is employed as the reference. 
However, directly measuring the pixel-wise difference between the predicted halftone and the reference does not work, since per-pixel inspection can never capture the intrinsic feature of binary halftone patterns.

\emph{Halftone Pattern Measurement}.
Inspired by perceptual loss\cite{zhang2018unreasonable}, we propose to measure the halftone pattern difference in the continuous feature domain. 
We pretrained an inverse halftoning network $\mathrm{F}$, a U-shaped architecture with three downscale blocks, four residual blocks, and three upscale blocks, to capture the halftone patterns in the continuous feature domain.   
Accordingly, we formulate the guidance loss $\mathcal{L}_{G}$ as
\begin{equation}
    \mathcal{L}_{G} = || \mathrm{F}(O_h) - \mathrm{F}(I_h) ||_2
\end{equation}
Then, we perform the warm-up training on the dithering network with the combined loss: 
\begin{equation}
    \mathcal{L}_{stage1} = \mathcal{L}_{half} + \mathcal{L}_{G}
\end{equation}
where we set $\alpha = 0.1$, $\beta = 0.6$, $\gamma = 0.3$.
The red curve in \figurename~\ref{fig:loss-curve} shows the high training efficiency. 
With only 28 epochs, it is able to generate decent visual results, as shown in \figurename~\ref{fig:guidance_loss}(c). 

In the second stage, we froze the predictor module; and trained the encoder and decoder networks to learn the desired halftone pattern.
The whole model was trained under the following combination of loss functions until the loss converged
\begin{equation}
    \mathcal{L}_{stage2} = \mathcal{L}_{half} + \mathcal{L}_{restore} + \epsilon \cdot \mathcal{L}_G
\end{equation}
By isolating the predictor module in training, we ensure that the learning of the encoder does not involve luminance information; only chrominance information is encoded into the halftone.
we set $\alpha = 0.4$, $\beta = 0.6$, $\gamma = 0.9$, $\epsilon = 0.3$ , $\zeta = 1$ and $\eta = 0.00002$ empirically.
It is worth noting that we set $\gamma$ as 0.9 instead of 0.3 as \oursbase. 
The detailed analysis and reasoning are discussed in Section \ref{sec:ablation}.
We still have to use guidance loss $\mathcal{L}_{G}$ here; as we experimented, if we dropped this loss, the halftone loses its structures and becomes over-smoothed.
\figurename~\ref{fig:guidance_loss}(d) shows an example of training without guidance loss in stage two.

At the final stage, we fine-tuned the predictor module. 
We adopted the inverse halftone module from \cite{xia2018deep} as our predictor, and it was trained by the loss function specified in \eqref{eq:lumin_loss} only.
The encoder and decoder are frozen in this stage.

The approach that separates stage two and stage three ensures that the encoder only encodes chrominance information into $O_h$.
The decoder only outputs two channels, compared to \oursbase, which is three.
Hence, the restoration burden on luminance has been shifted to the predictor.
Such modification allows the encoder to generate $O_h$ with better blue-noise quality instead of sacrificing it.
Therefore, our proposed method generates halftones with better tone resemblance and blue-noise quality, while maintaining its restoration quality compared to our base design \oursbase.

\section{Experimental Results}
\label{sec:exp_result}


We trained the warm-up stage with 28 epochs until the model generated decent visual results, then we trained the second stage and final stage until both corresponding losses converged. Each stage takes 87 epochs and 50 epochs respectively. The whole training takes a total of 165 epochs to complete.
It is obvious that our predictor embedded method contains more parameters on the restoration side than our base method \oursbase.
Therefore, to further justify the effectiveness of the predictor module, we compare our method with different variations of \oursbase~in this section.
\note{color_map}{
The color error maps, in this paper and the supplementary, are generated by normalizing the pixel from [0,255] to [0,1] and computing the L1 distance between the images in the RGB color space.
}

\subsection{Dataset}

We evaluated our method on the VOC2012 dataset\cite{my-pascal-voc-2012}. 
It contains 17,125 color images. 
We cropped and resized all images into $256\times256$.
We randomly split the image set into:
    13,758 images as training set; and 
    3,367 images as validation set.


\subsection{Comparison with traditional halftoning}
\label{sec:exp_result:halftone}
\begin{figure}[!t]
    \centering

    \includegraphics[width=\linewidth]{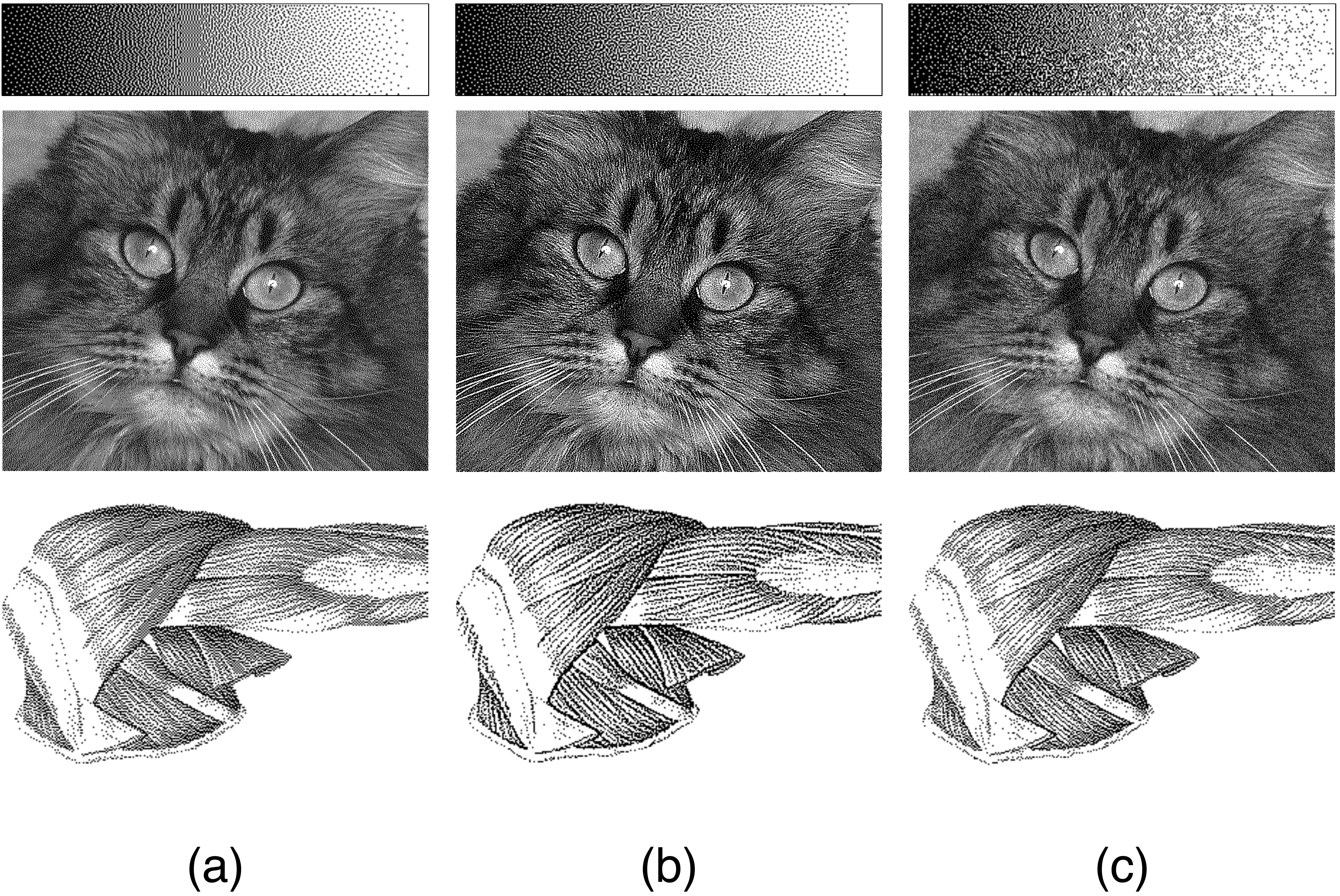}
    \caption{Qualitative comparison of halftone images with intensive structures.
    (a) Ostromoukhov;
    (b) Structure-aware; and
    (c) Ours.
    }
    \label{fig:halftone_classic}
\end{figure}

\begin{figure}[!t]
    \centering

    \includegraphics[width=\linewidth]{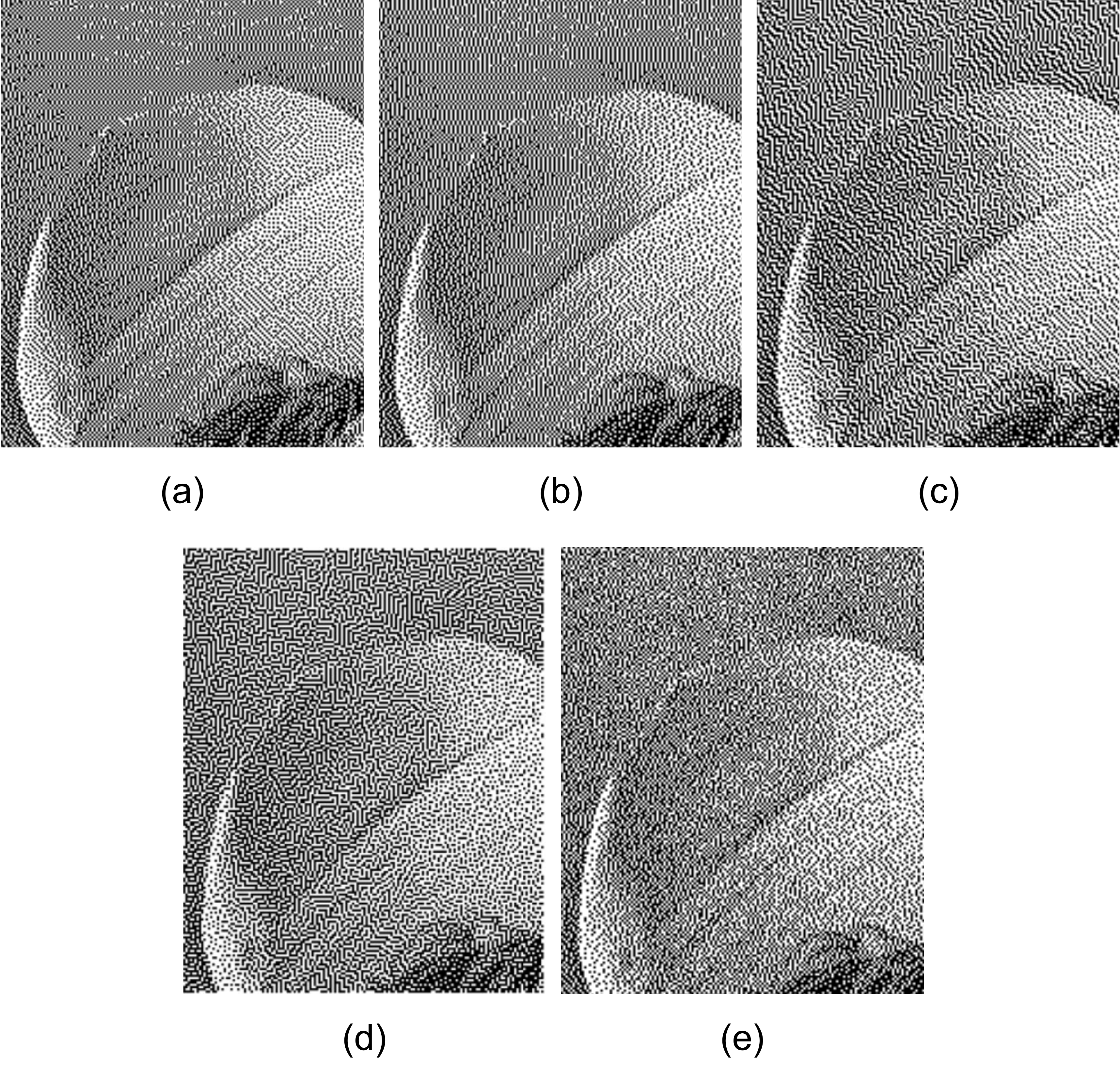}
    \caption{
    Qualitative comparison of halftone patterns on ``Lena''.
    The example of TDED$_\text{BS}$ is directly from \cite{fung2016tone}.
    (a) Floyd-Steinberg;
    (b) Ostromoukhov;
    (c) Zhou and Fang \cite{zhou2003improving};
    (d) TDED$_\text{BS}$ \cite{fung2016tone}; and
    (e) Ours.
    }
    \label{fig:halftone_lena}
\end{figure}

\begin{table}[!t]
\renewcommand{\arraystretch}{1.3}
\caption{Quantitative evaluation on halftone images in terms of the mean PSNR and SSIM values. Higher PSNR/SSIM indicate better quality.}
\label{tab:halftones}
\centering
\small
\begin{tabular}{c | c c }
\hline
\bfseries Methods & \bfseries PSNR & \bfseries SSIM \\
\hline
Ostromoukhov method         & 41.728    & 0.1007 \\
Structure-aware halftoning  & 21.803    & 0.0340 \\
Ours                        & 34.444    & 0.1094 \\
\hline
\end{tabular}
\end{table} 
Following the practice in \cite{pang2008structure}, the tone consistency is measured by PSNR between the Gaussian-filtered halftone and the Gaussian-filtered luminance channel of the input, and the structure consistency is measured by SSIM between the halftone and the luminance channel of the input.
We experimented with 3,367 grayscale images (decolorized from our validation set), as existing halftoning methods can only dither grayscale images. 
Two classical halftoning methods that generate high-quality halftones are selected as our competitors, Ostromoukhov's method \cite{ostromoukhov2001simple} and the structure-aware halftoning method \cite{pang2008structure}. 
In our experiment, the structure-aware halftoning method is used with default parameters for quantitative evaluation while the case-by-case tuned result is provided for visual comparison. 
The statistics are tabulated in Table \ref{tab:halftones}. 
Among all, our method achieves the best comprehensive performance of tone similarity (PSRN) and structure similarity (SSIM). 
\figurename~\ref{fig:halftone_classic} shows examples on a gray ramp and images with structures.
Our halftone resembles the continuous tone but is not as smooth as the traditional methods. 
It is because we traded off the blue-noise quality for encoded color information.
However, our halftone visual quality is comparable with traditional methods in images with structures. 
Our method achieves better structure than the error-diffusion method \cite{ostromoukhov2001simple}, and less rigid patterns compared to the structure-aware method \cite{pang2008structure}.
We further compared our method with some state-of-the-art halftoning methods \cite{zhou2003improving, fung2016tone}. 
\figurename~\ref{fig:halftone_lena} shows that our method produce less ``worm effects'' than \cite{ostromoukhov2001simple}, \cite{pang2008structure} but still produce checkboard patterns compared to those improved methods \cite{zhou2003improving, fung2016tone}.

\begin{figure}[!t]
    \centering

    \includegraphics[width=\linewidth]{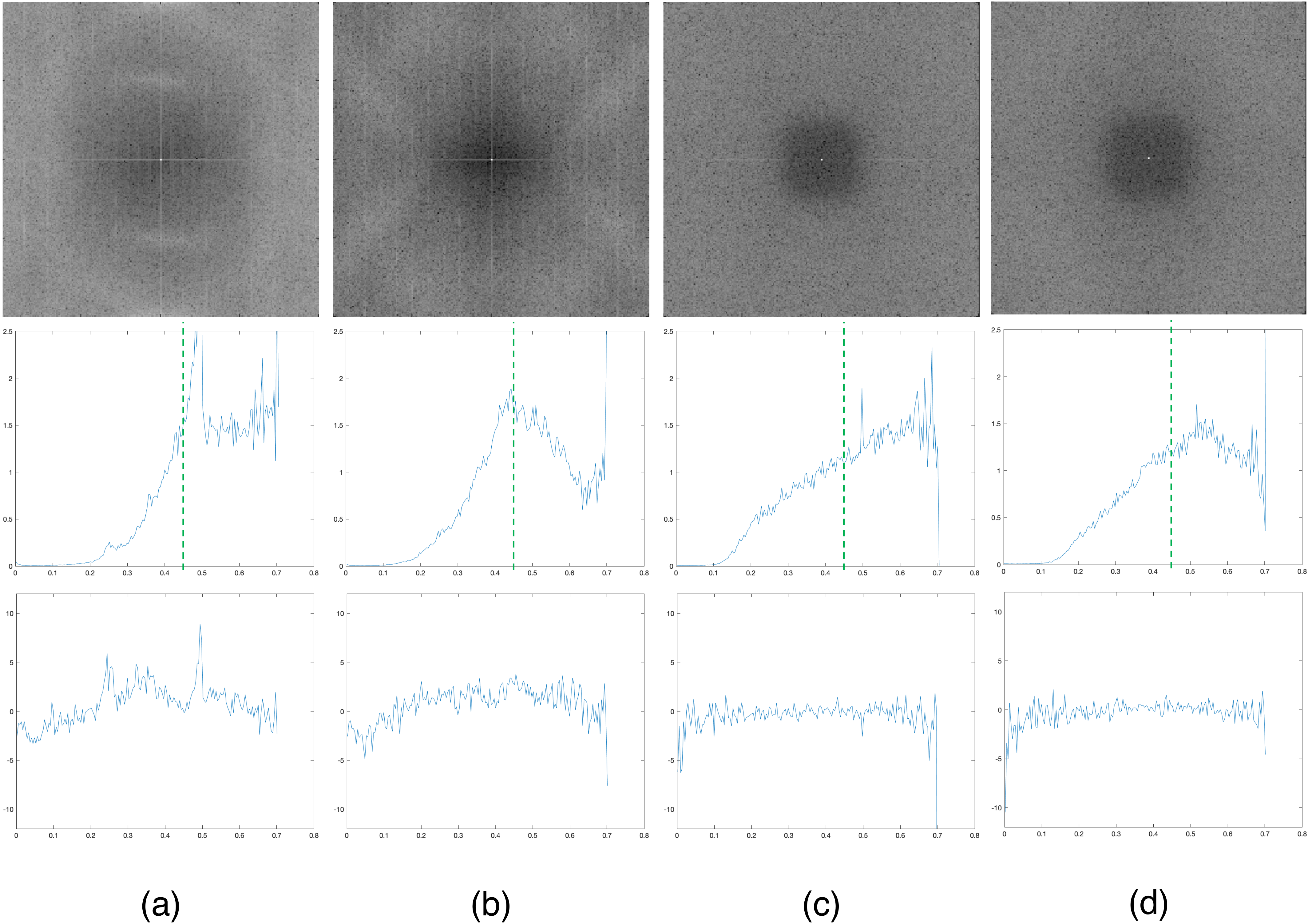}
    \caption{
    Spectrum analysis on various halftone results from constant-grayness 0.8. From top to bottom: the power spectrum, radially averaged power spectrum density and anisotropy.
    The green dashed line indicates the principle frequency.
    (a) Floyd-Steinberg;
    (b) Ostromoukhov;
    (c) Ours/base; and
    (d) Ours
    }
    \label{fig:specturms}
\end{figure}




%
To analyze the blue-noise quality of our halftone images, we adopted the common analysis methods as in \cite{ulichney1988dithering}.
We selected the classical error diffusion methods\cite{floyd1976adaptive, ostromoukhov2001simple} as our competitors.
We analyzed halftone images obtained from constant-grayness images in terms of its \emph{power spectrum} and its \emph{radially averaged power spectrum}.
The grayness is set to 0.8.
The power spectrum indicates the frequency amplitude in 2-D.
Since the amplitude of frequency in halftone is supposed to be radially symmetry.
The radially averaged power spectrum visualizes the 2-D power spectrum in 1-D space.
According to \cite{ulichney1988dithering}, for a good halftone with blue-noise property, the radial frequency graph should have 
1) low amplitude in low-frequency areas; 
2) a peak transition region on principle frequency; 
3) a flat high-frequency region.
We adopted the principle frequency defined in \cite{ulichney1987digital}.
\figurename~\ref{fig:specturms} illustrates the power spectrum and radial averaged power spectrum of the converted halftone image.
Our method produces low amplitude in low-frequency regions similar to \cite{floyd1976adaptive, ostromoukhov2001simple}.
Also, we observed our peak is closer to the principle frequency, and the shape of the curve resembles the shape in the classical method \cite{ostromoukhov2001simple}. The frequency analysis among different gray levels is located in \figurename~\ref{fig:supp_spectrum} in the supplementary.


\subsection{Evaluation on Reversible Halftoning}
\begin{figure*}[!t]
    \centering
    \includegraphics[width=\linewidth]{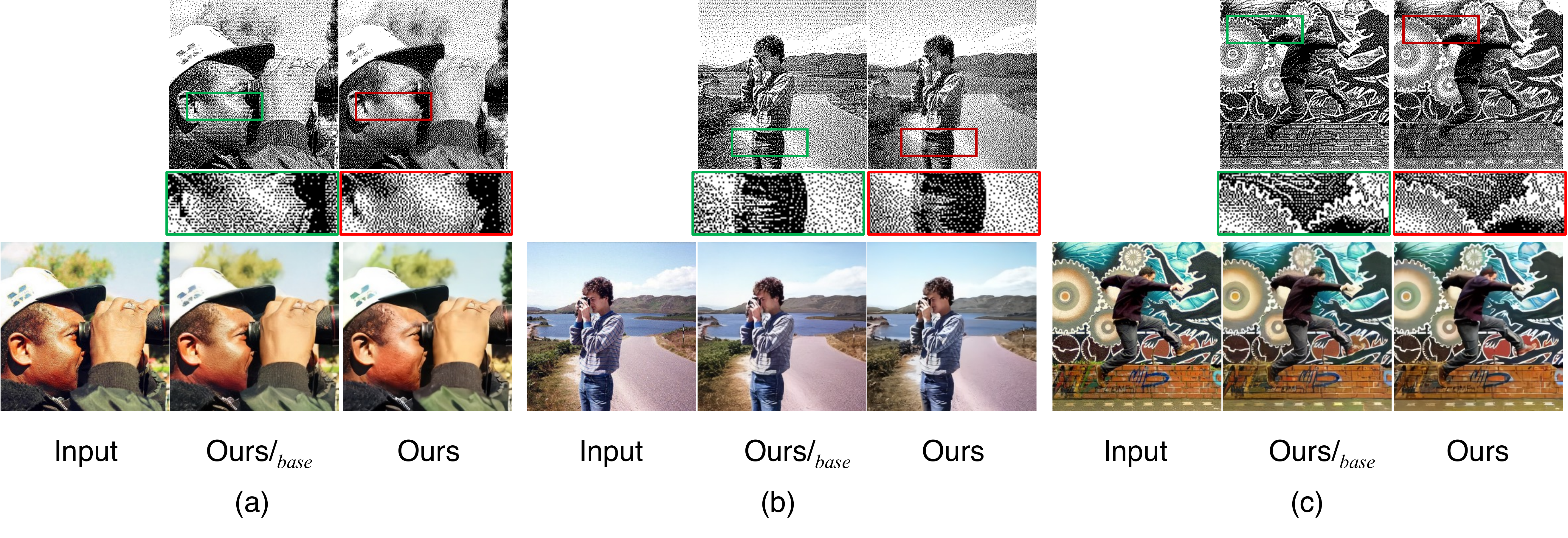}
    \caption{
    Qualitative comparison on halftone image and restored image together.
}
    \label{fig:halftone-and-restored-compare}
\end{figure*}

\begin{figure}[!t]
    \centering
    \includegraphics[width=\linewidth]{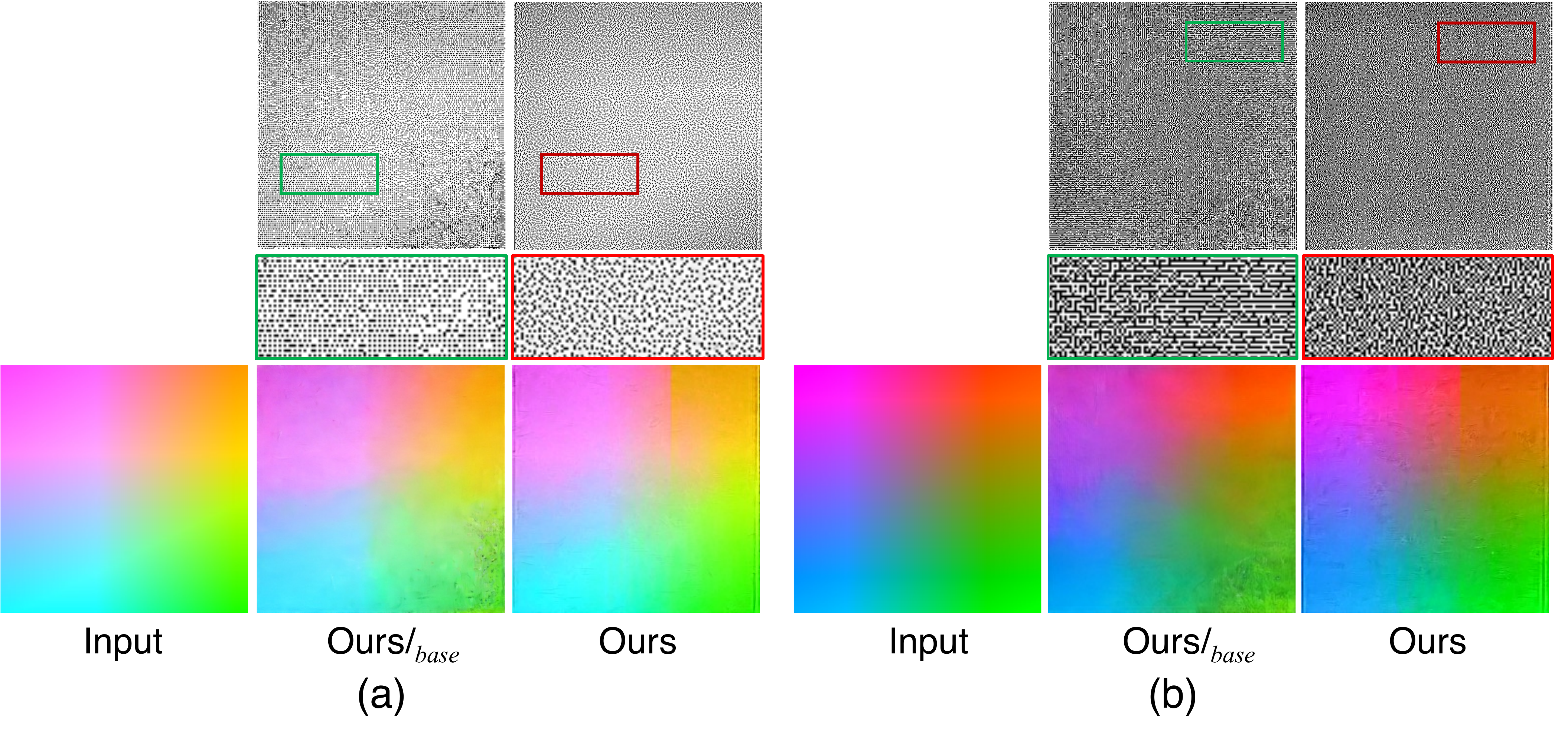}
    \caption{
    Qualitative comparison on halftone image and restored image together on color ramp. (a) Y=0.8; and (b) Y=0.5.
}
    \label{fig:halftone-and-restored-compare-colorramp}
\end{figure}

\noindent \textbf{Blue-noise quality} \quad
We evaluate the blue-noise quality on the halftone generated by our method, which includes the predictor module, and our base method \oursbase~over color input images.
\figurename~\ref{fig:halftone-and-restored-compare} (top-row) shows halftone examples produced by \oursbase~and our method.
Both methods preserve the structural details.
Our halftone patterns produce smoother surfaces and less "grid-like" structures in low-variance areas.
This indicates a better blue-noise property on our halftones.
\note{explain}{
The improvement of blue-noise quality is much more evident on \figurename~\ref{fig:halftone-and-restored-compare-colorramp}(a). 
Our halftone dissolves the "grid-like" patterns and is visually smoother than \oursbase with comparable restoration quality.
More examples of the color ramp are provided in \figurename~\ref{fig:supp_color_ramp} in the supplementary. 
}

By observing the spectrum analysis results on \oursbase~and our method in \figurename~\ref{fig:specturms}, we can see that our halftone resembles the transition peak closer to \cite{ostromoukhov2001simple} than \oursbase~in power spectrum.
Although the peak region is not as wide as \cite{ostromoukhov2001simple}; and shifted right from the principle frequency, it approaches the principle frequency closer than our base methods.
Hence, our method with a predictor module extends the model's ability to produce halftone images with better blue-noise quality. 

\begin{table}[!t]
\renewcommand{\arraystretch}{1.3}
\caption{Quantitative evaluation on halftone and restored color images in terms of the mean PSNR and SSIM values. Higher PSNR/SSIM indicate better quality.}
\label{tab:restore}
\centering
\small
\resizebox{\linewidth}{!}{%
\begin{tabular}{c | c c | c c }
\hline
                  & \multicolumn{2}{c|}{\bfseries Halftone} & \multicolumn{2}{c}{ \bfseries Restoration} \\
\bfseries Methods & \bfseries PSNR  & \bfseries SSIM        & \bfseries PSNR    & \bfseries SSIM         \\
\hline
 PRL-net\cite{xia2018deep} + ColTran\cite{kumar2021colorization} & 41.728 & 0.1007 & 19.543 & 0.4612 \\
\oursbase & 30.624 & 0.1440  & 28.130    & 0.8592 \\
Ours      & 32.283 & 0.1136 & 27.292    & 0.6060 \\
\hline
\end{tabular}
}
\end{table}

\begin{figure}[!t]
\centering
\includegraphics[width=\linewidth]{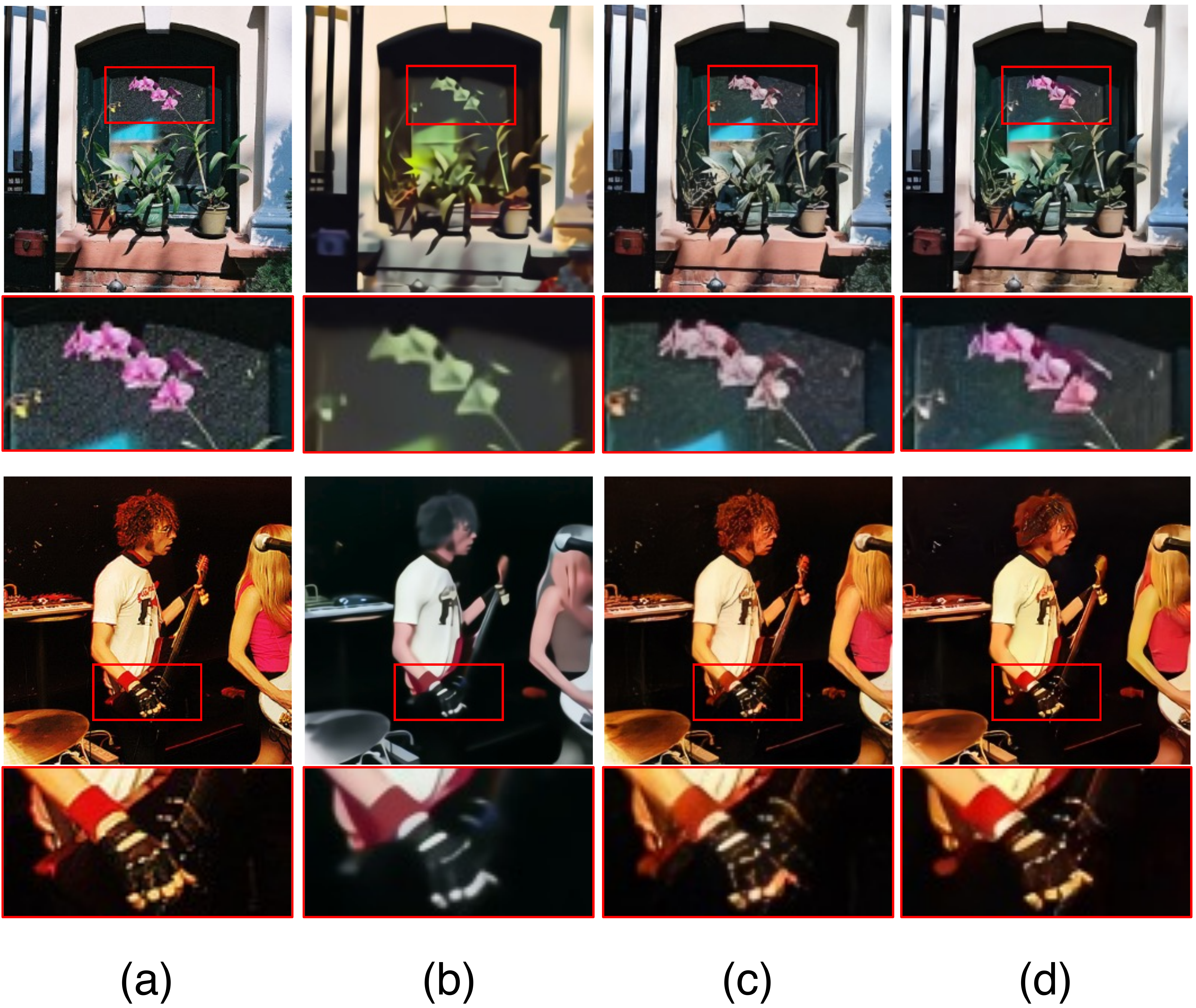}
\caption{Colorization vs. our methods.
(a) Input;
(b) ColTran\cite{kumar2021colorization};
(c) \oursbase; and
(d) Ours
}
\label{fig:restore_coltran}
\end{figure}

\noindent \textbf{Restoration quality} \quad
We compare our method in grayscale and color image inputs.
We take two state-of-the-art methods as our competitors: the PRL-Net\cite{xia2018deep} as our baseline grayscale candidates and the ColTran\cite{kumar2021colorization} as baseline color candidates.
The PRL-Net\cite{xia2018deep} generates grayscale from the error-diffused halftone, and ColTran\cite{kumar2021colorization} colorize the grayscale from \cite{xia2018deep} to obtain the color version.
Since PRL-Net can only restore grayscale images, we prepared 3,367 grayscale images (decolorized from our testing dataset) for grayscale comparison.
Table \ref{tab:restore} presents the statistics of both PSNR and SSIM.
Our superiority lies in the restoration of the color domain.
Our method avoids the ill-posed problem of color choice in areas and improves color segmentation with encoded color information.
ColTran\cite{kumar2021colorization} experiences the drop in PSNR due to differences in color choice from the ground truth.
\figurename~\ref{fig:restore_coltran} shows an example of ColTran\cite{kumar2021colorization} failing to segment color section properly. 
Our method is able to produce the pink and green color at corresponding areas which ColTran\cite{kumar2021colorization} failed. 
In fact, the ability of ColTran\cite{kumar2021colorization} to guess color lies in its training batches while our method retrieves the color information from the halftone patterns.

Furthermore, we compare our method with \oursbase~to evaluate the effectiveness of the predictor module.
The example in \figurename~\ref{fig:halftone-and-restored-compare} (bottom-row) demonstrates our improved encoded halftone with comparable restoration ability with \oursbase.
By adopting the predictor module, we achieve the same level of restoration quality while improving the blue-noise quality in halftone.
It is because, as our halftone becomes smoother with less encoded information, the predictor module fills in the missing luminance information by "guessing". Therefore, our restoration power maintains a comparable level with \oursbase~when the "guess" is correct.
\note{explain2}{
We notice the restoration artifacts in extreme dark luminance, such as Y=1 in Figure \ref{fig:supp_color_ramp}. 
We believe it is caused by the inverse halftone module being trained on images with structural complexity, rather than plain colors. 
Nonetheless, our restoration quality is comparable on average and applicable in real-world cases.
}

\subsection{Data embedding study}

We adopt the concept of entropy in information theory to estimate encoded information in our halftone patterns.
The information of a source produced can be measured in terms of entropy\cite{baronchelli2005measuring}.
We claim that our method encoded less information than \oursbase. 
One way to evaluate the estimated entropy is to compare signals via a lossless compression.
In lossless compression, redundant signals are replaced by shorter code words\cite{shannon1948mathematical}.
Therefore, sources with less information, hence less "surprise" in signals, should obtain a higher compression rate\cite{baronchelli2005measuring, avinery2019universal}. 

Since compressing images with different spatial arrangements yield variances in compression rates, We rotated and flipped all 3,367 images in all four directions before evaluating their compression rates. By expanding the image set in this way, we could ensure that the compression rates we compared were representative of each image's general compression characteristics. Finally, we compressed all the halftone images into a single ZIP archive using the universal zip library\cite{gailly2004zlib} for comparison purposes.
Table \ref{tab:zipper} shows the respective compression rates.
The classical error-diffusion method obtains the highest compression ratio, while all variations of our base methods obtain lower compression rates. 
Our predictor-embedded method sits between the error-diffusion method and our base methods.
This experiment further proves our claim.

\begin{table}[!t]
\renewcommand{\arraystretch}{1.3}
\caption{Entropy estimations via lossless universal zip compression.}
\label{tab:zipper}
\centering
\small
\begin{tabular}{c | c}
\hline
\bfseries Methods & \bfseries Compression rate \\
\hline


Ostromoukhov's\cite{ostromoukhov2001simple}     & 87.0\% \\
\oursbase                                       & 86.2\% \\
\oursbaseBplus                                  & 86.3\% \\
\oursbaselarge                                  & 86.6\% \\
\oursbaselargeBplus                             & 86.6\% \\
Ours                                            & 86.8\% \\

\hline
\end{tabular}
\end{table}



\begin{figure}[!t]
\centering
\includegraphics[width=\linewidth]{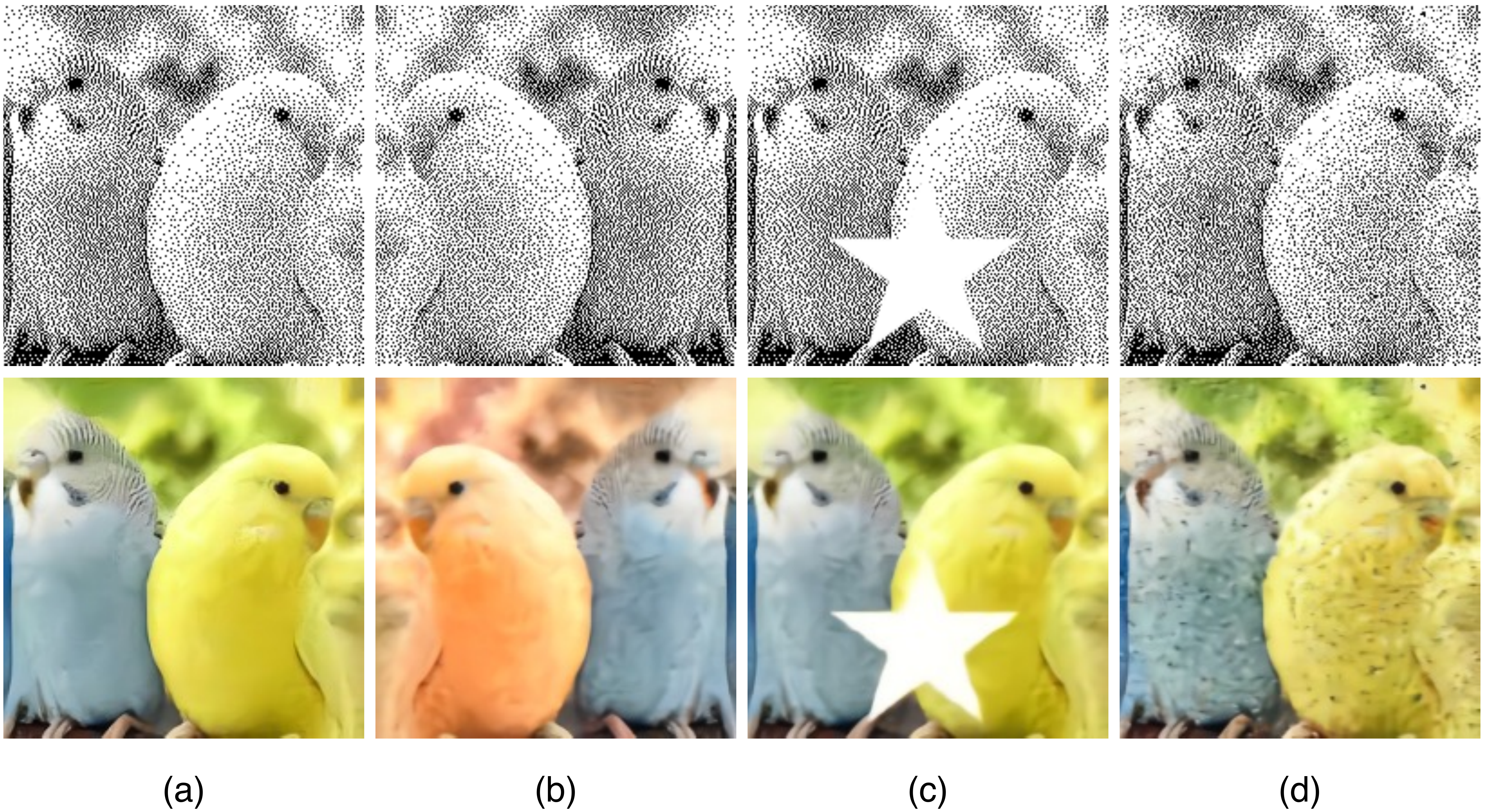}
\caption{Robustness study of reversible halftones.
The color image (bottom row) is restored from the reversible halftones (top row).
(a) No operation;
(b) flipped;
(c) partial masked; and
(d) random noise with 10\% impulse noise, which is more destructive than Gaussian noise.
}
\label{fig:robustness}
\end{figure}

\begin{figure}[!t]
\centering
\includegraphics[width=\linewidth]{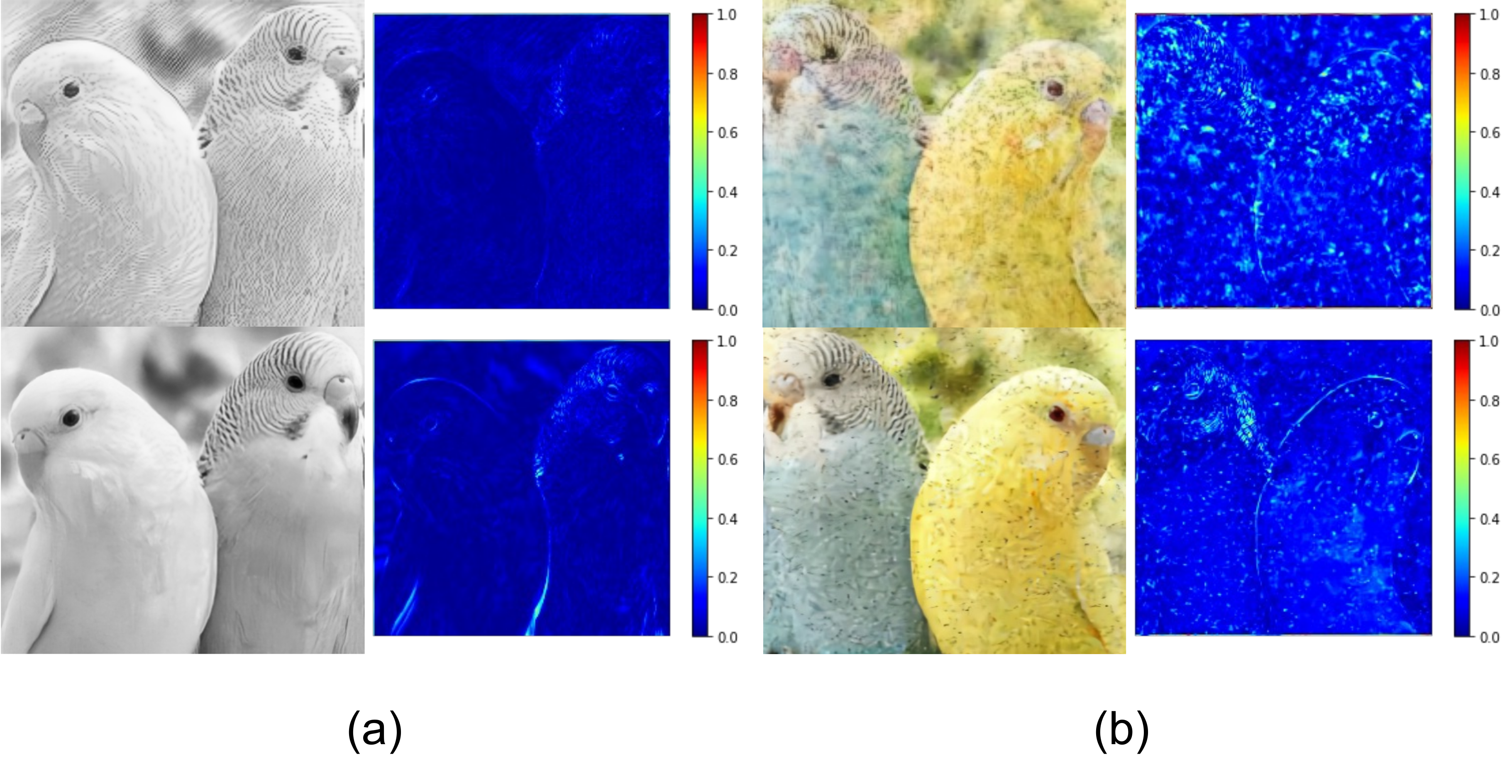}
\caption{Robustness comparison between flipping and random noise. The top is \oursbase~and the bottom is Ours.
(a) Flipped (grayscale); and 
(b) random noise.
}
\label{fig:robustness_vs_reshalf}
\end{figure}

Furthermore, we analyze how the encoding information is embedded and its robustness by applying several typical disturbances to the generated halftones, including flipping, partial removal, and random impulse noises.
\figurename~\ref{fig:robustness} illustrates the restored color examples from augmented halftones.
\figurename~\ref{fig:robustness}(c) shows when under a regional mask, the color in the unmasked regions is restored similarly to the original. This indicates our color data are encoded local-wise instead of global-wise. However, the restored structure is also blurred. We believe this is brought by the prediction accuracy of the predictor.
The incorrect color restored from the flipped halftone in \figurename~\ref{fig:robustness}(b) indicated the encoded information is directionally sensitive.
Although both \oursbase~and Ours cannot restore a correct color from the flipped halftones, our restored version contains fewer structural diagonal artifacts. \figurename~\ref{fig:robustness_vs_reshalf}(a) shows a comparison of the grayscale version of the restored image. 
\figurename~\ref{fig:robustness_vs_reshalf}(b) further shows that our method increases the tolerance to noise against our base method, which indicates the good potential to be used in real-world applications.
Since most of the structural information is constructed from the luminance and we offload such work to the predictor, the encoded information only affects the color correctness.
Therefore, our restored color in flipped halftones contains fewer structural artifacts than \oursbase.
Our method also shows higher tolerance to random noises than \oursbase.
Ground-truth images and detail comparisons are located in \figurename~\ref{fig:supp_robustness_vs} in the supplementary.

\subsection{Ablation Study}
\label{sec:ablation}

In this section, we demonstrate the necessity of the fine-tuning stage of the predictor.
In the training iteration, we froze the predictor so it does not participate in the gradient backward propagation.
Ours/$_{p-froze}$ in Table \ref{tab:ablation} shows a huge drop in restoration accuracy.
\figurename~\ref{fig:invhalf_freeze} in the supplementary shows an example of this case.
Since we adopted the pretrained model from \cite{xia2018deep}, which trained on classic error-diffusion halftone images, as our predictor, the pretrained model cannot recognize our halftone pattern and treat them as noise (\figurename~\ref{fig:invhalf_freeze}(c)).
Therefore, the predicted luminance contains artifacts.

\begin{table}[!t]
\renewcommand{\arraystretch}{1.4}
\caption{Ablation study on various training methods.}
\label{tab:ablation}
\centering
\small
\begin{tabular}{c | c c | c c}
\hline
                        & \multicolumn{2}{c|}{\bfseries Halftone} & \multicolumn{2}{c}{ \bfseries Restoration} \\
\bfseries Methods       & \bfseries PSNR    & \bfseries SSIM    & \bfseries PSNR    & \bfseries SSIM \\
\hline
\oursbase                   & 30.625           & 0.1459            & 27.822           & 0.6462 \\
Ours/$_{p-froze}$   & 32.360           & 0.1148            & 24.115           & 0.4835 \\
Ours/$_{end-to-end}$      & 30.854           & 0.1467            & 28.995           & 0.6850 \\
Ours                        & 32.680           & 0.1152            & 27.686           & 0.6565 \\
\hline
\end{tabular}
\end{table}
We studied the importance of the isolation training strategy.
We released the predictor and trained all three modules end-to-end in the second stage, skipping stage three.
The result is denoted as Ours/${_{end-to-end}}$ in Table \ref{tab:ablation}.
We can see that there is an improvement in color restoration.
However, the halftone accuracy remains the same level as \oursbase.
It is because, in the backward propagation stage, both the predictor and decoder act as the learning factors for the encoder.
With the increased parameters on the restoration side, an improvement in color restoration is expected.
Therefore, isolating the predictor when training the encoder becomes a necessary step.
This indicates the significance of our two-stage strategy.

\begin{table}[!t]
\renewcommand{\arraystretch}{1.4}
\caption{Ablation study on variation of $\mathcal{L}_{blue}$'s coefficient $\gamma$.}
\label{tab:ablation_loss_blue}
\centering
\small
\begin{tabular}{c | c | c c | c c}
\hline
                        &   & \multicolumn{2}{c|}{\bfseries Halftone} & \multicolumn{2}{c}{ \bfseries Restoration} \\
\bfseries Methods       & $\gamma$  & \bfseries PSNR    & \bfseries SSIM    & \bfseries PSNR    & \bfseries SSIM \\
\hline
\multirow[c]{2}{*}{\oursbase}           & 0.3       & 30.678        & 0.1426            & 28.234        & 0.6408 \\
                                        & 0.9       & 30.124        & 0.1838            & 27.666       & 0.6524 \\
\hline
\multirow[c]{2}{*}{\oursbaselarge}      & 0.3       & 29.927        & 0.2422            & 28.276       & 0.6331 \\
                                        & 0.9       & 28.744        & 0.1536            & 21.353       & 0.4489 \\
\hline
\multirow[c]{2}{*}{Ours}                & 0.3       & 36.194        & 0.1081            & 29.362       & 0.7025 \\
                                        & 0.9       & 32.282        & 0.1133            & 27.451       & 0.6036 \\
\hline
\end{tabular}
\end{table}

\begin{figure}[!t] 
	\centering
	\includegraphics[width=\linewidth]{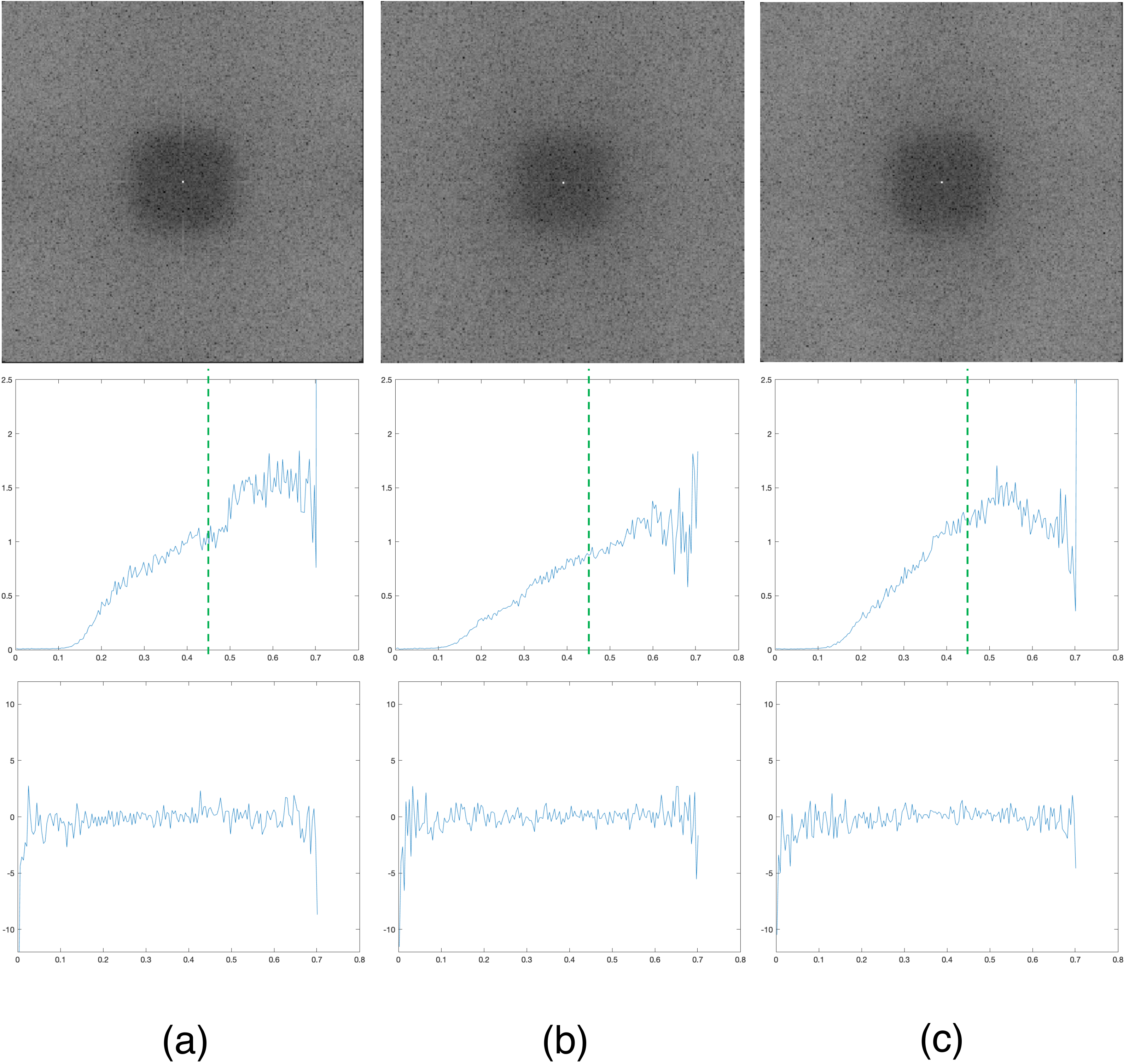}
	\caption{
 Spectrum analysis between \oursbase~and Ours with increased blue-noise loss weight.
 (a) \oursbase, $\gamma=0.9$~;
 (b) \oursbaselarge, $\gamma=0.9$~; and
 (c) Ours.
 }
    \label{fig:specturm_ablation}
\end{figure}

Finally, we compare the effectiveness of blue-noise manipulation between our base and predictor-embedded methods.
For a fair comparison, we trained our base method with doubled layers in the decoder module, denoted as \oursbaselarge, to match the parameter size with the predictor-embedded approach.
It is worth noting that we propose setting the coefficient $\gamma = 0.3$ in our base method because we found that $\mathcal{L}_{blue}$ and $\mathcal{L}_{restore}$ are conflicting each other.
However, with the predictor-embedded approach, we can push the value $\gamma$ to 0.9.
Table \ref{tab:ablation_loss_blue} shows the detailed comparison between the variations of $\gamma$ in training.
We can see that if we increase the weight of blue-noise loss, \oursbase~results in the same level of quality regarding halftone accuracy and restoration accuracy. 
\figurename~\ref{fig:specturm_ablation} shows the spectrum analysis with models trained with increased $\gamma$.
Even with a larger parameter size in the decoder and higher blue-noise loss weight on \oursbase, the anisotropy was suppressed, but the model still struggles to produce a transition peak around the principle frequency in the power spectrum.
We can see that both \figurename~\ref{fig:specturm_ablation}(a) and \figurename~\ref{fig:specturm_ablation}(b) produce high intensity in the high-frequency area.
It is because without changing the encoding content and the suppression of low frequency introduced by the blue-noise loss $\mathcal{L}_{blue}$, the model is forced to encode information into the high-frequency area.
\oursbase~reaches its limits to improve the blue-noise property.

\begin{figure}[!t] 
	\centering
	\includegraphics[width=\linewidth]{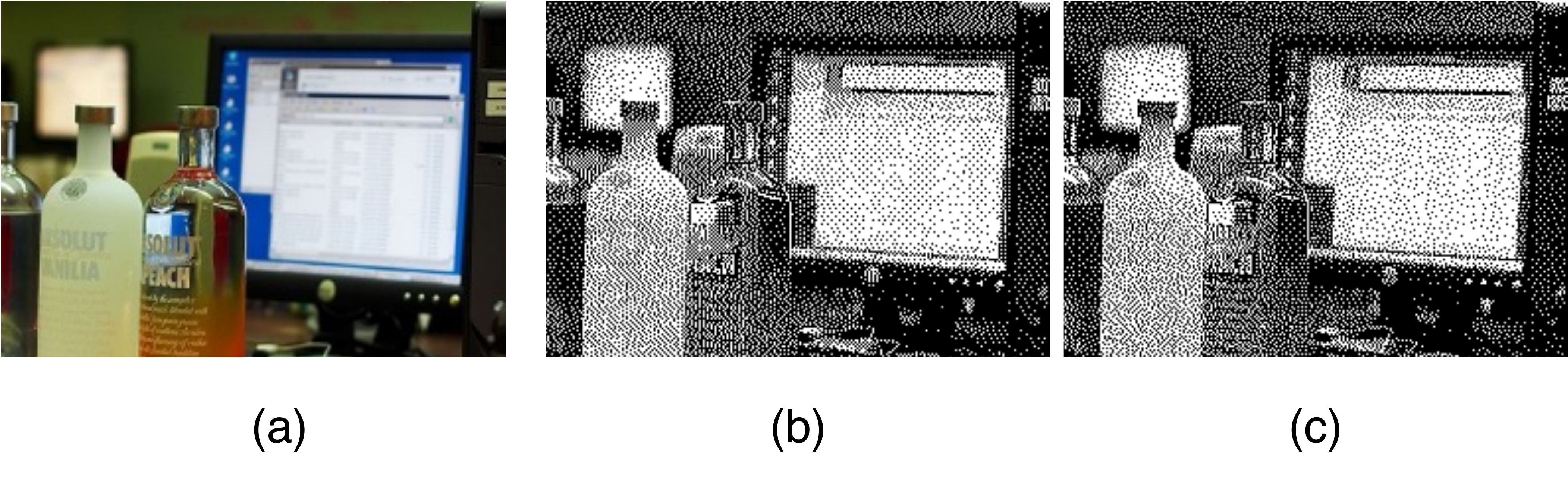}
	\caption{
 Example of halftone generated via different tone and blue-noise loss weights.
 (a) Input
 (b) $\beta=0.6,\gamma=0.3$; and
 (c) $\beta=0.6,\gamma=0.9$
 }
 \label{fig:tone_vs_blue}
\end{figure}

With the predictor approach, our model quickly raises the bar of halftone tone consistency and restoration accuracy with the default weighting of 0.3.
It is because, with lesser information that needs to be encoded, the model tends to improve the tone of halftone patterns when we set the default $\mathcal{L}_{tone}$'s coefficient $\beta = 0.6$.
Since we aim to improve the blue-noise quality in our halftone images, we take $\gamma=0.9$ as our final proposed version.
\figurename~\ref{fig:tone_vs_blue} shows an example of halftone images between heavier tone consistency vs. heavier blue-noise weights.

The ablation study of the NIB block can be found in the supplementary.

\section{Conclusion}
\label{sec:conclusion}

We propose a novel reversible halftoning technique with high restoration ability and state-of-the-art visual quality. Our approach is a strong alternative to traditional halftoning methods and eliminates the need to tackle the ill-posed inverse halftoning problem. To extend the ability of the reversible model, we introduce a predictive module that offloads the encoding burden between the blue-noise property and the hidden color information. Our formulation of the blue-noise loss as a low-frequency constraint on constant-grayness guarantees the visual pleasantness of halftone patterns. We also propose a method to modulate the priorities of different loss terms in three stages to handle the tricky optimization landscape. Our experiments demonstrate the advantages of our approach and highlight the improvement achieved by the predictor strategy. We believe our contributions to reversible halftoning and the predictor approach will inspire future work in this field.

\ifCLASSOPTIONcaptionsoff
  \newpage
\fi

\bibliographystyle{IEEEtran}
\bibliography{database}

%

\begin{IEEEbiography}[{\includegraphics[width=1in,height=1.25in,clip,keepaspectratio]{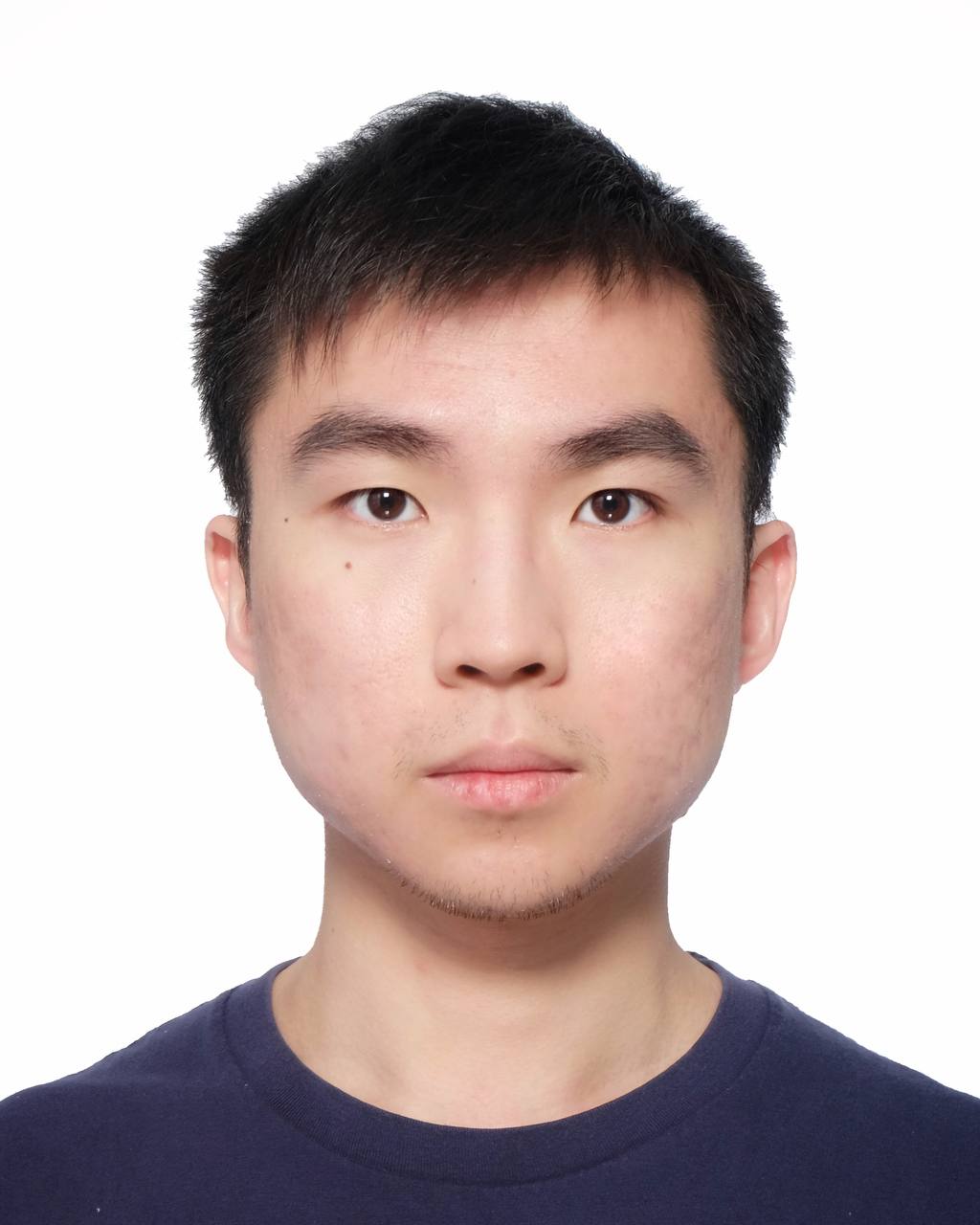}}]{Cheuk-Kit Lau} received a B.Sc degree in Computer Science from The Chinese University of Hong Kong in 2017. He begins to pursue an M.Phil degree with the Department of Computer Science \& Engineering of the Chinese University of Hong Kong in 2021. His research interests include image processing, video generation, deep learning and computer vision.
\end{IEEEbiography}
\begin{IEEEbiography}[{\includegraphics[width=1in,height=1.25in,clip,keepaspectratio]{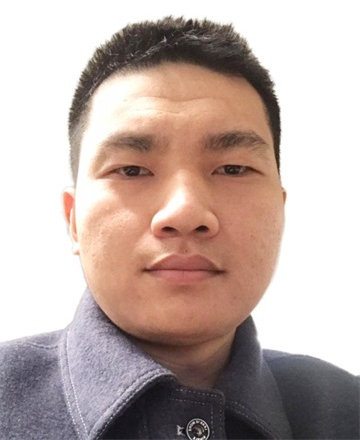}}]{Menghan Xia}
received a B.Eng. degree in Remote Sensing Science and Techniques in 2014 and a Master degree in Pattern Recognition and Intelligent Systems in 2017 from Wuhan University. In 2021, he obatained his Ph.D. degree in Computer Science from The Chinese University of Hong Kong. He is currently a senior researcher in Tencent AI Lab. His research interests include image \& video generation, facial animation, cross-modal translation and deep learning.
\end{IEEEbiography}
\begin{IEEEbiography}[{\includegraphics[width=1in,height=1.25in,clip,keepaspectratio]{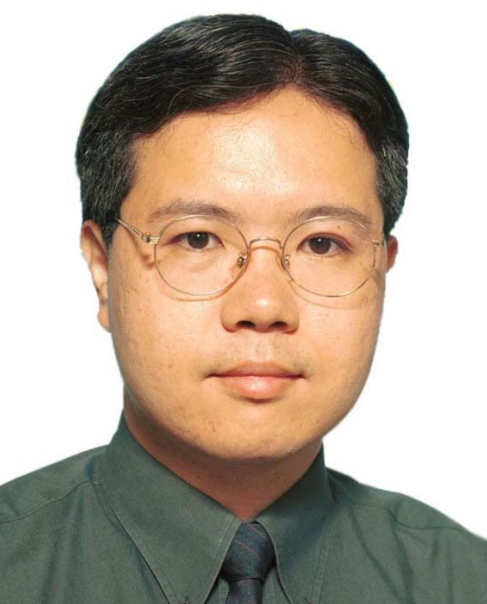}}]{Tien-Tsin Wong}
graduated from the Chinese University of Hong Kong in 1992 with a B.Sc. degree in Computer Science. He obtained his M.Phil. and Ph.D. degrees in Computer Science from the same university in 1994 and 1998, respectively. In August 1999, he joined the Computer Science \& Engineering Department of the Chinese University of Hong Kong. He is currently a professor. He is a core member of the Virtual Reality, Visualization and Imaging Research Centre in The Chinese University of Hong Kong. His main research interests include computer graphics, computational manga, precomputed lighting, image-based rendering, GPU techniques, medical visualization, multimedia compression, and computer vision.
\end{IEEEbiography}

\vfill

\clearpage




\begin{figure*}[!t]
    \centering
    \includegraphics[width=\linewidth]{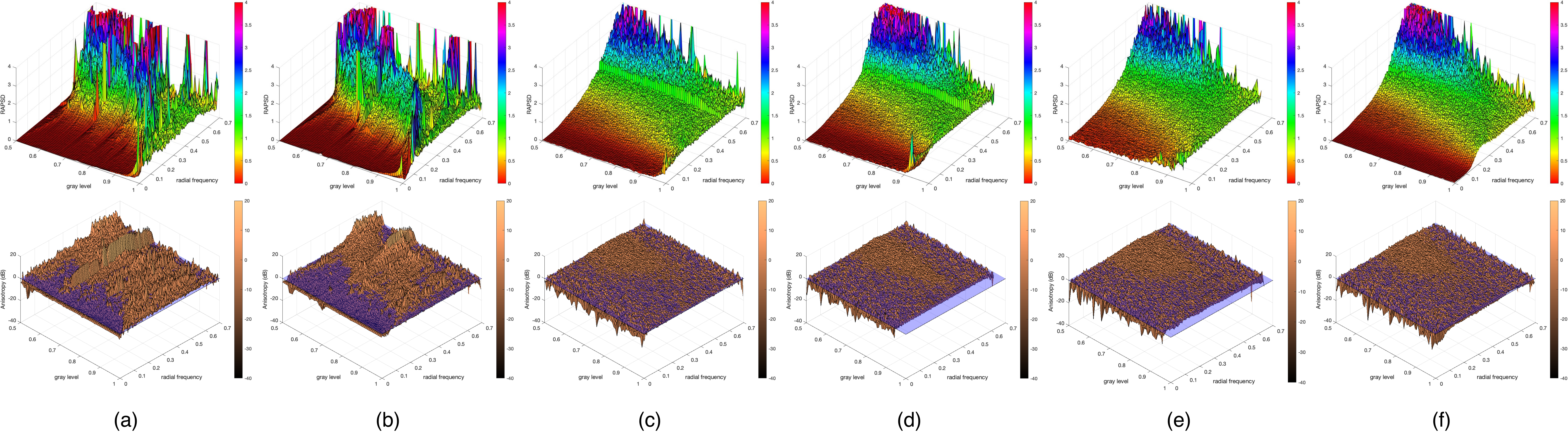}
    \caption{Specturm analysis on different gray levels.
    From top to bottom: radially averaged power spectrum density and anisotropy. 
    (a) Floyd-Steinberg;
    (b) Ostromoukhov;
    (c) \oursbase;
    (d) \oursbaseBplus;
    (e) \oursbaselargeBplus; and
    (f) Ours.
    }
    \label{fig:supp_spectrum}
\end{figure*}

\begin{figure}[!h]
    \centering
    \includegraphics[width=0.48\textwidth]{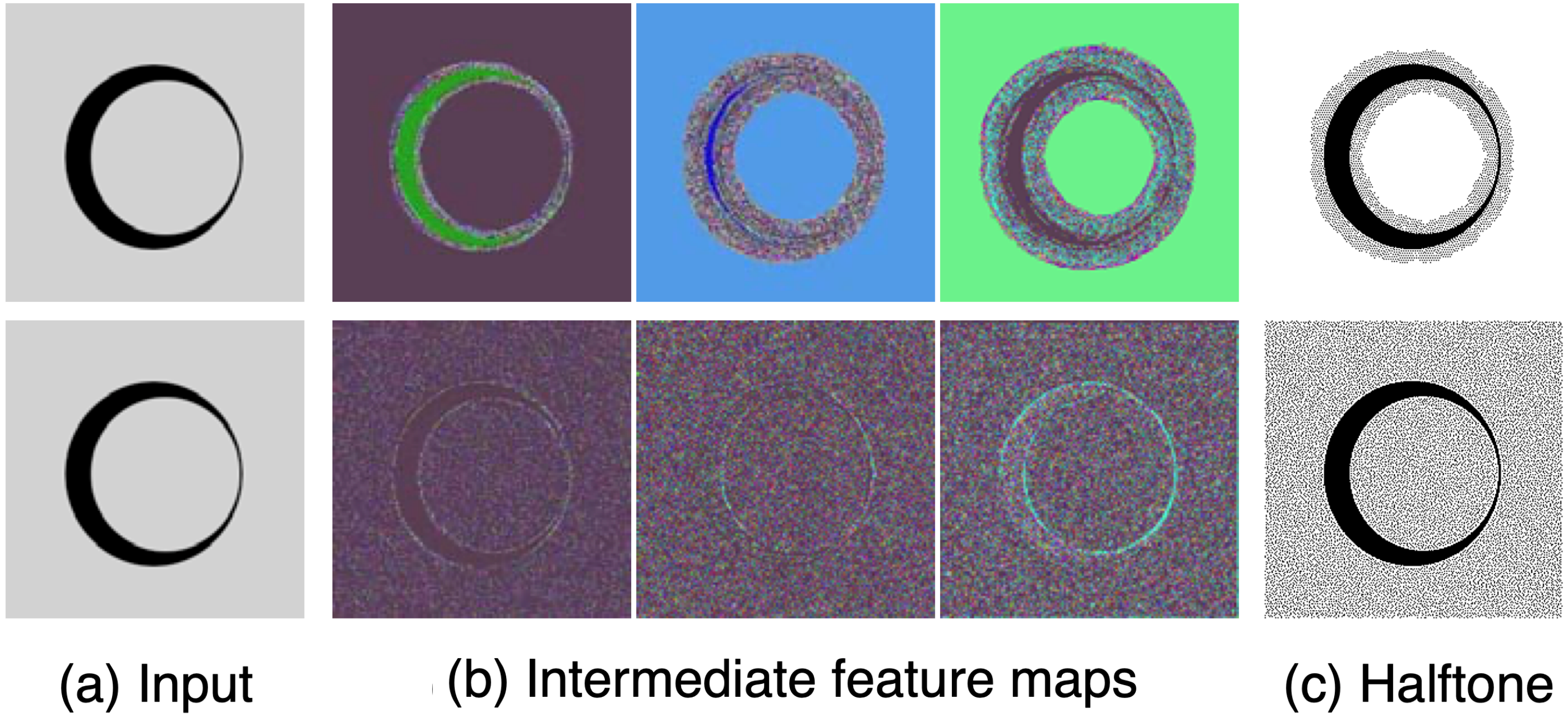}
    \caption{Visualization of CNN halftoning. Due to the flatness degradation, typical CNNs fail to generate spatial variation in flat regions (up row); The NIB equipped CNNs can address the limitation effectively (bottom row).}
    \label{fig:degradation}
\end{figure}

\begin{figure}[!h] 
	\centering
	\includegraphics[width=\linewidth]{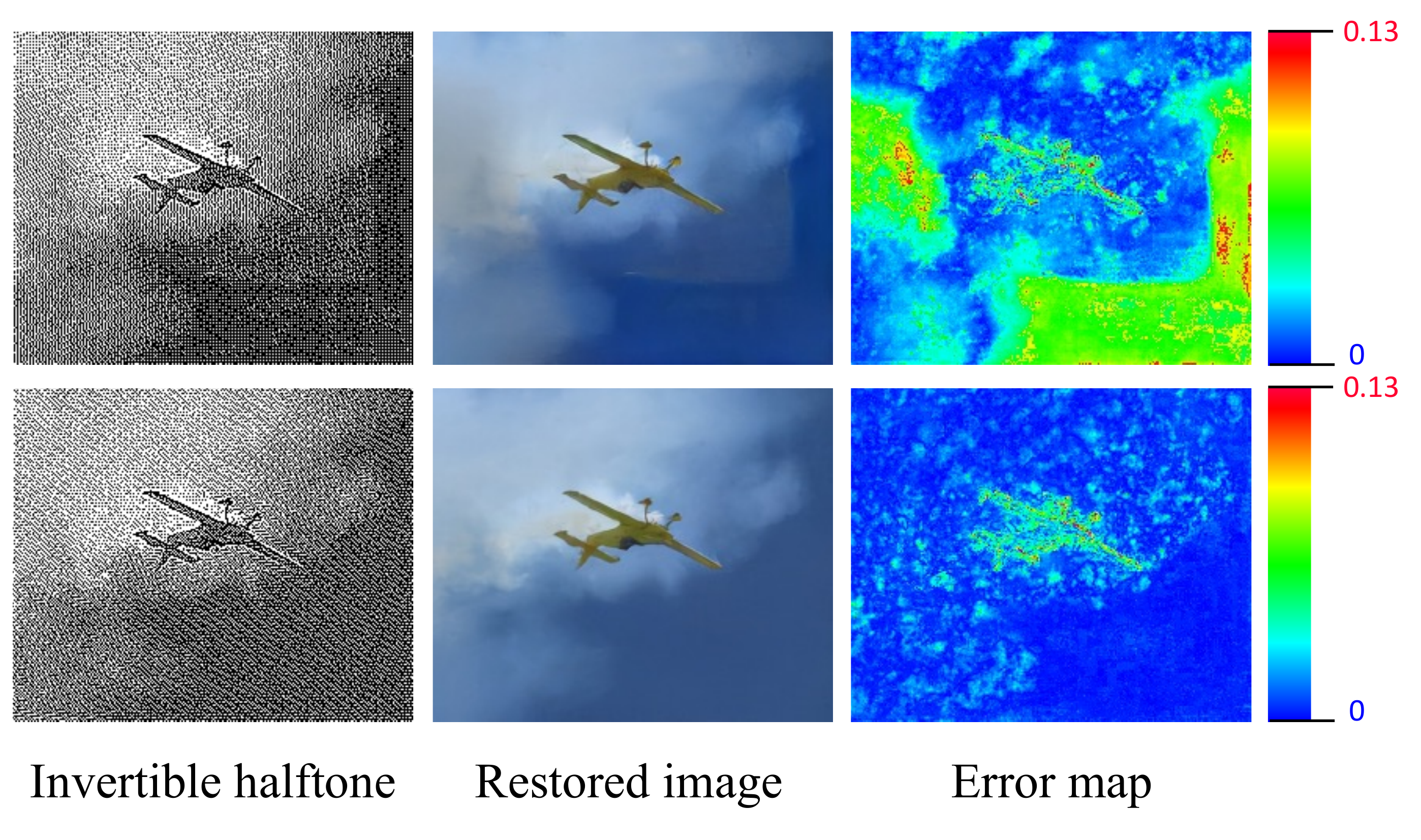}
	\caption{
 Performance comparison between the model without (top row) and with NIB equipped (bottom row). The color-coded error maps visualize the deviation w.r.t the ground truth.}
 	\label{fig:nib_noise_feed}
\end{figure}

\begin{table}[!h]
\renewcommand{\arraystretch}{1.4}
\caption{Ablation analysis on noise incentive block (NIB). Statistic over the color testing dataset.}
\label{tab:nib_ablation}
\centering
\small
\resizebox{\linewidth}{!}{%
\begin{tabular}{c|c|cc|cc}
		\hline
		\multirow{2}{*}{\bfseries Category}      &\multirow{2}{*}{\bfseries Variant}      & \multicolumn{2}{c}{\bfseries PSNR}       & \multicolumn{2}{c}{\bfseries SSIM}   \\
		\cline{3-6}                                 &                                                 & \bfseries Mean           & \bfseries Stddev                 & \bfseries Mean   			& \bfseries Stddev	  	  \\ \hline
		\multirow{2}{*}{Halftoning}    & Ours/NIB                                & 31.915		 & 1.8185     	  	     & 0.1514		    & 0.0827		 \\
        \cline{2-6}                                & Ours                                       & 33.734		   & 0.6078     	  	   & 0.1702		      & 0.0906	       \\ \hline
        \multirow{2}{*}{Restoration} &  Ours/NIB                               & 27.743	        & 2.2795               & 0.8667            & 0.0420		    \\
        \cline{2-6}                                 & Ours                                       & 29.112	       & 2.9705                & 0.8826            & 0.0430       \\ \hline
	\end{tabular}
}
\end{table}

\begin{figure}[!h]
    \centering
    \includegraphics[width=\linewidth]{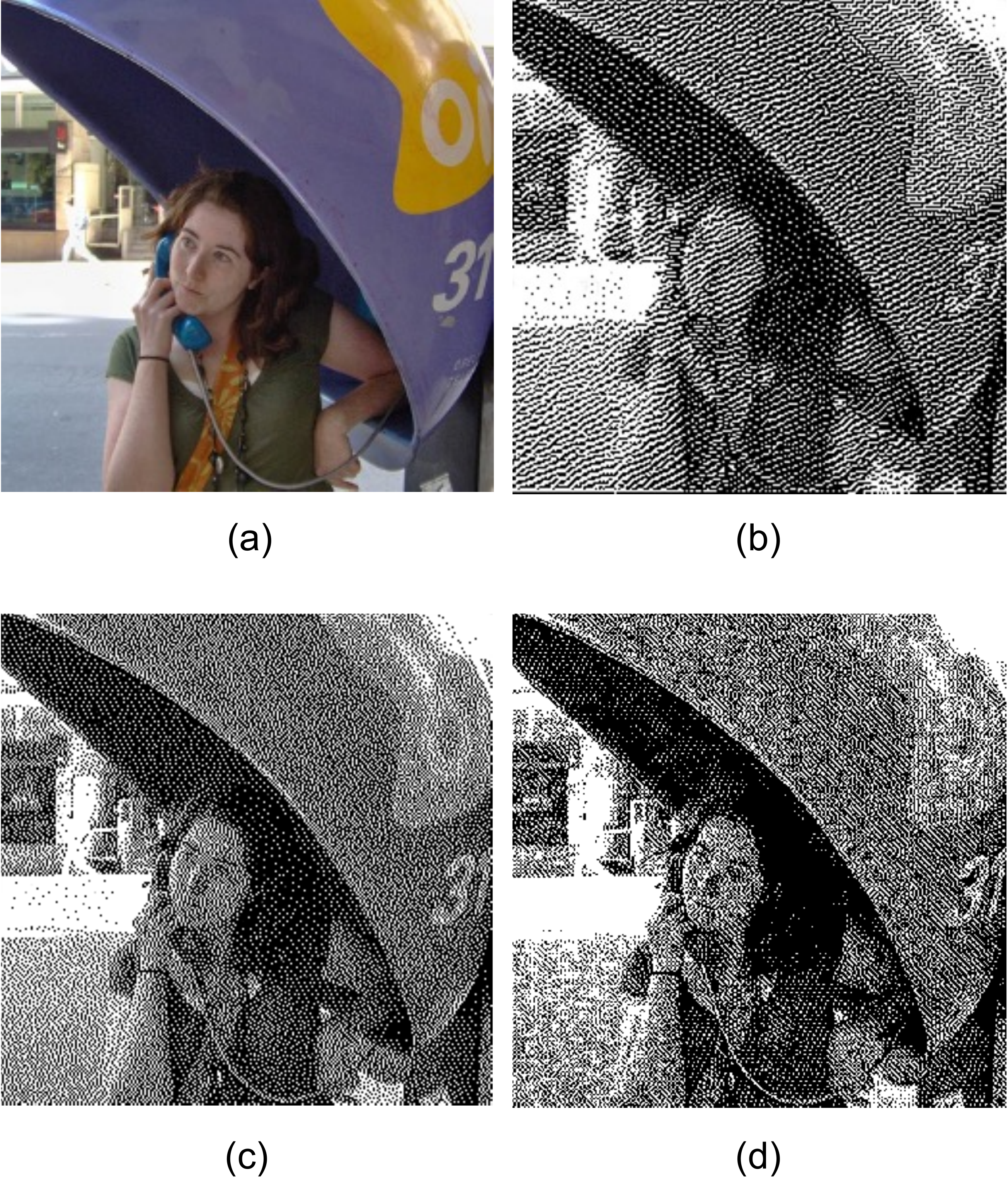}
    \caption{Visual comparison of the weight of guidance loss in the second stage.
    (a) Input;
    (b) $\epsilon = 0$
    (c) $\epsilon = 0.3$; and
    (d) $\epsilon = 1$.
    }
    \label{fig:supp_guidance_loss}
\end{figure}

\noindent \textbf{Ablation Study of Noise Incentive Block} \quad
As mentioned in Section \ref{subsec:incentive_block}, our proposed noise incentive block (NIB) enables the dithering network to generate binary halftones for constant input.
To further analyze the effect, we conduct an ablation study on the NIB of our dithering network. 
Note that the blue-noise loss cannot be applied when NIB is not used since it is formulated on the dithered constant-grayness. 
Regarding this, we intentionally remove the blue-noise loss in all model variants to avoid inducing other factors. 

The quantitative result of the color testing dataset is given in Table \ref{tab:nib_ablation}.
The statistics show that equipping NIB to the dithering network improves halftone generation and color image restoration. It is probably because the randomness introduced by NIB favors the dithering process, i.e., focusing on pattern distribution instead of individual pixel values. 
In addition, CNNs also partially degrade in smooth regions, which hinders the generation of desired halftone patterns.
\figurename~\ref{fig:nib_noise_feed} shows an example to verify this hypothesis.

\begin{figure}[!h]
    \centering
    
    \subfloat[]{\includegraphics[width=.24\linewidth]{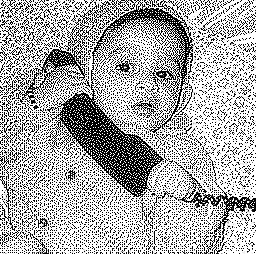}\label{fig:invhalf_freeze:halftone}}
    \hfil
    \subfloat[]{\includegraphics[width=.24\linewidth]{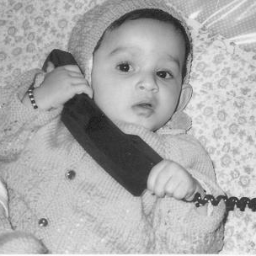}\label{fig:invhalf_freeze:gt}}
    \hfil
    \subfloat[]{\includegraphics[width=.24\linewidth]{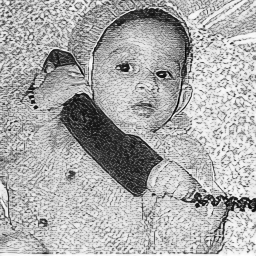}\label{fig:invhalf_freeze:pretrained}}
    \hfil
    \subfloat[]{\includegraphics[width=.24\linewidth]{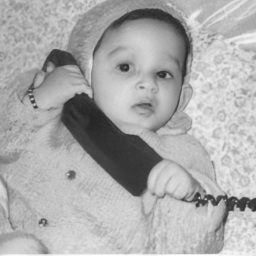}\label{fig:invhalf_freeze:finetune}}
    
    \caption{Example of luminance obtained from inverse halftone module\cite{xia2018deep}.
    (a) Our halftone
    (b) Luminance Ground-truth
    (c) Luminance via pretrained \cite{xia2018deep}
    (d) Luminance via fine-tuned \cite{xia2018deep}
    }
    \label{fig:invhalf_freeze}
\end{figure}

\begin{figure}[!h]
    \centering
    \includegraphics[width=\linewidth]{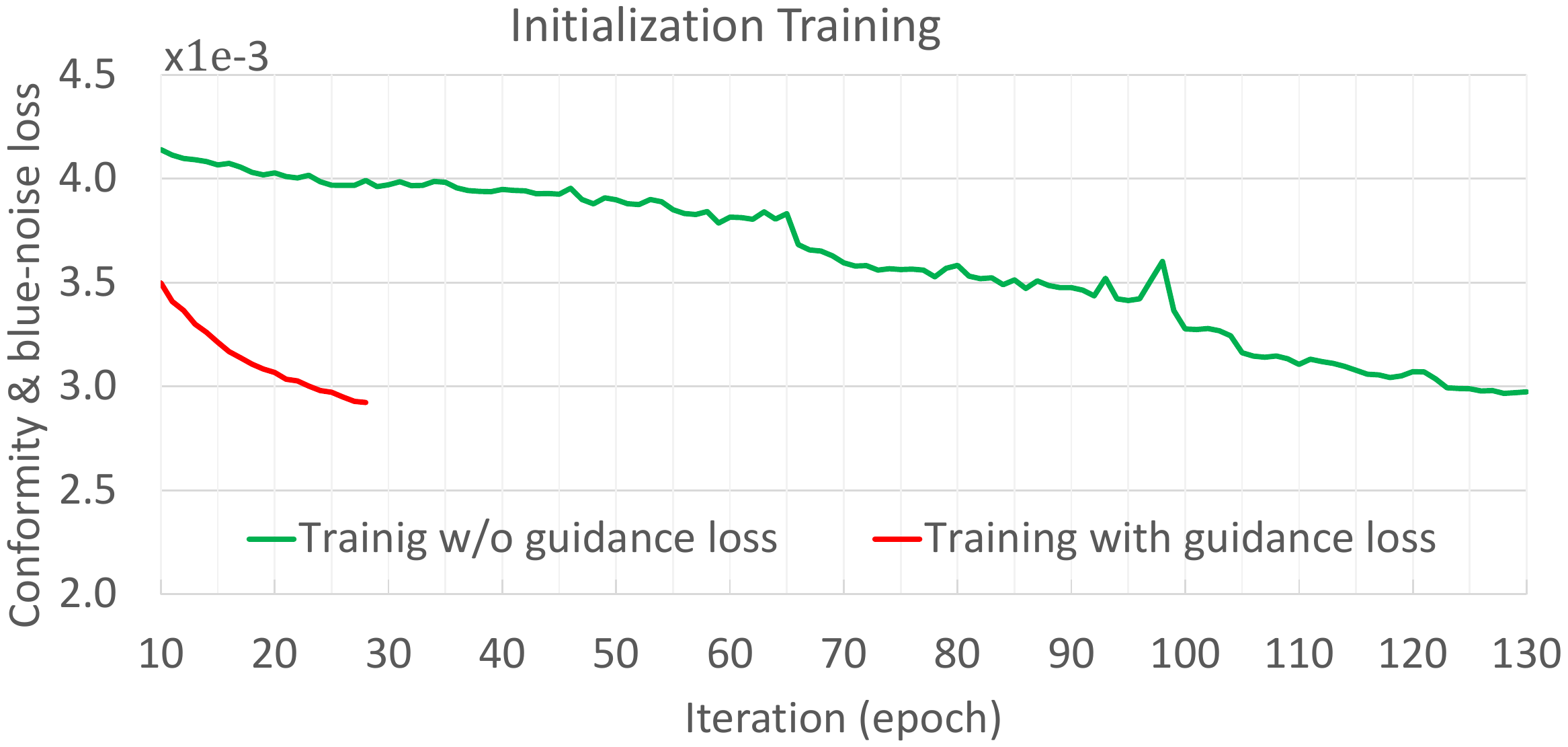}
    \caption{Halftone visual loss against iteration of warm-up training with and w/o the guidance loss $\mathcal{L}_{G}$}
    \label{fig:loss-curve}
\end{figure}

\begin{figure*}[!h]
    \centering
    \includegraphics[width=\linewidth]{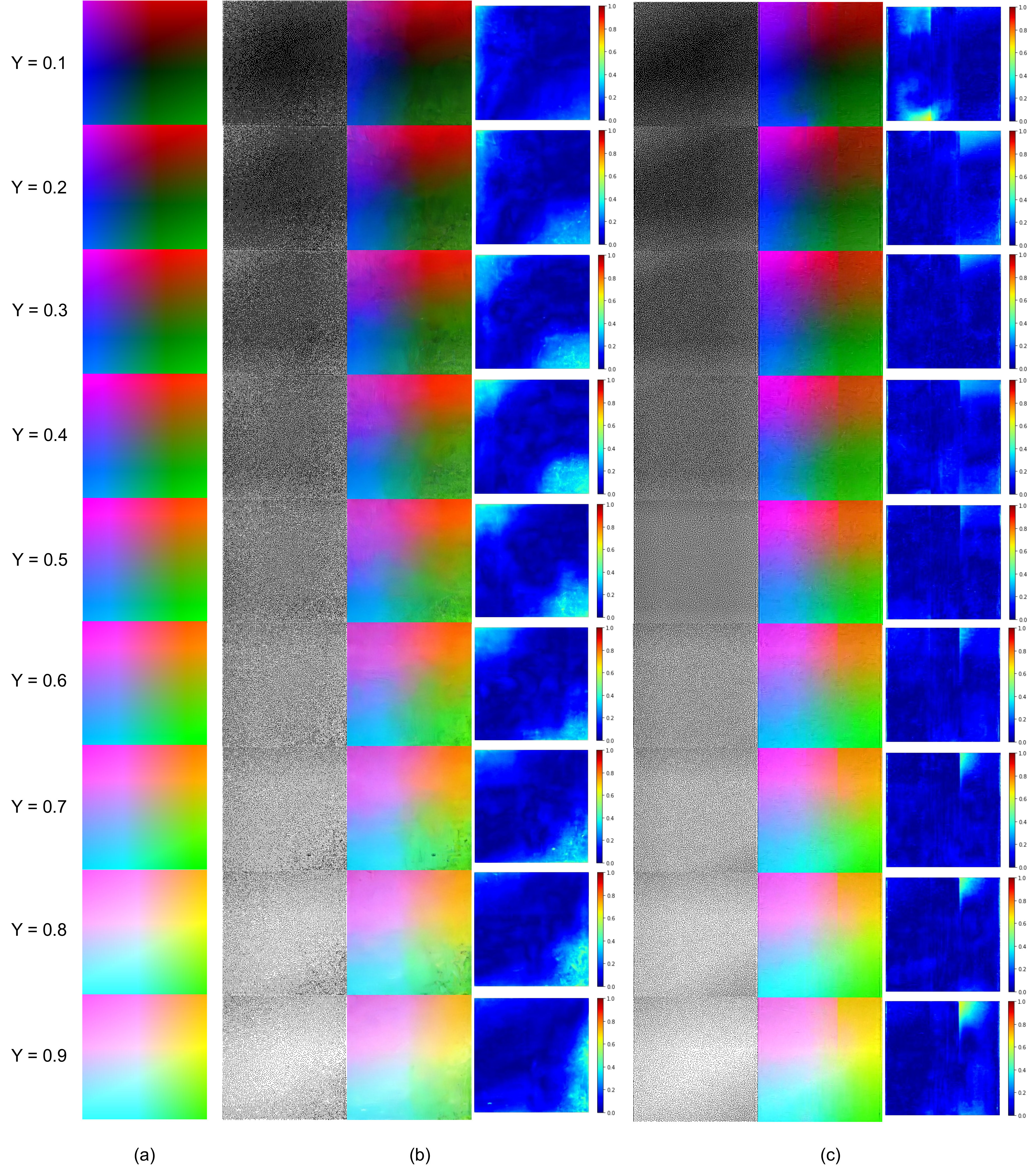}
    \caption{
    Qualitative comparison on halftone image and restored image together on color ramp. (a) Input; (b) \oursbase; and (c) Ours.
    }
    \label{fig:supp_color_ramp}
\end{figure*}

\begin{figure*}[!h]
    \centering
    \includegraphics[width=\linewidth]{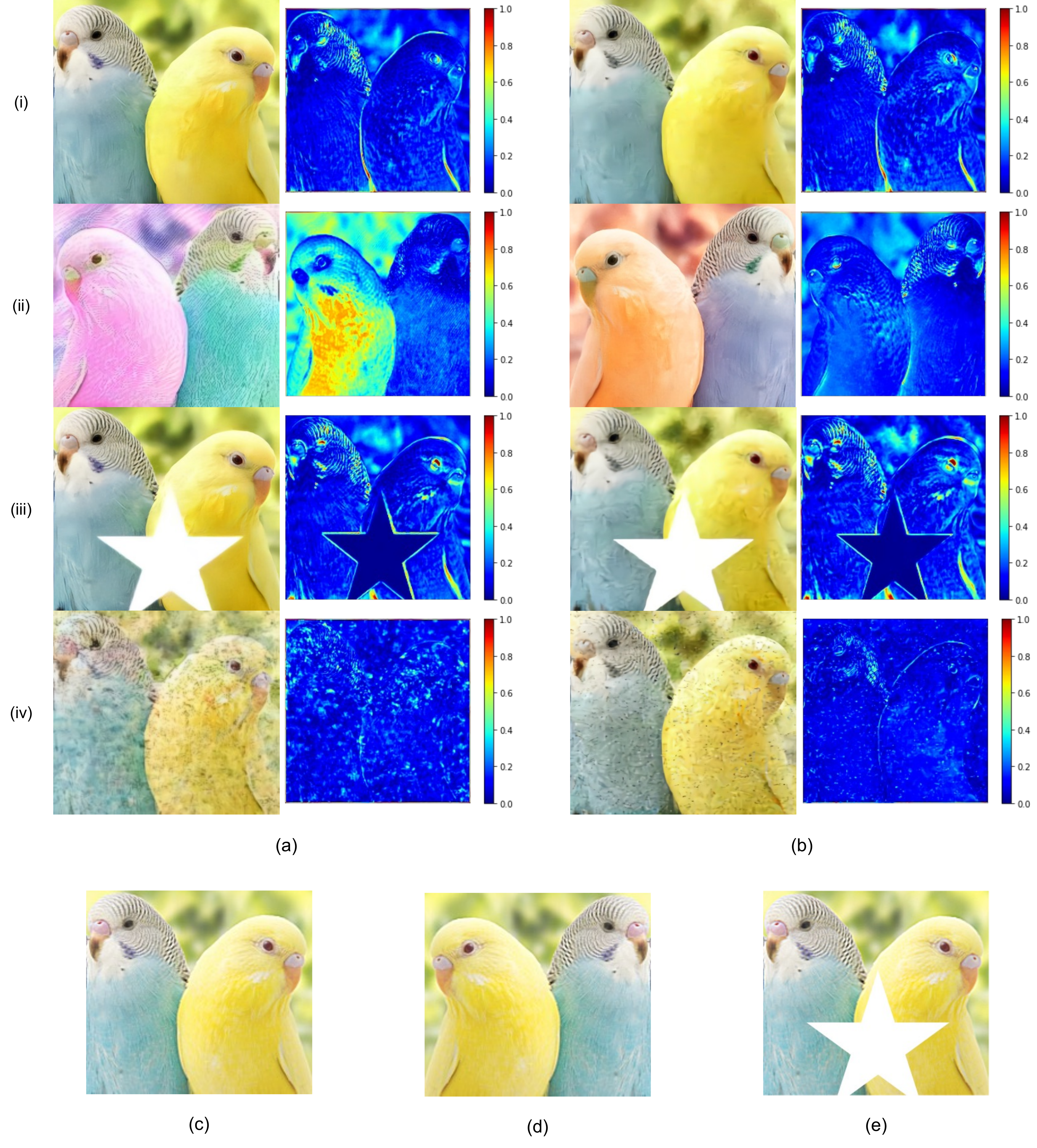}
    \caption{
    Robustness comparison between \oursbase~and Ours. 
    From top to bottom are color restored from (i) original; (ii) flipped; (iii) masked; and (iv) random noised reversible halftone. 
    For the ground-truth image for error map comparison, we compare (i), (iv) with (c); (ii) with (b); and (iii) with (c). 
    (a) \oursbase; (b) Ours; and (c), (d), (e) are ground-truth images.}
    \label{fig:supp_robustness_vs}
\end{figure*}



\end{document}